%% file: main_arxiv.tex
\documentclass[10pt,twocolumn,letterpaper]{article}

\usepackage{cvpr}
\usepackage{amsmath,amsfonts}
\usepackage{array}
\usepackage{textcomp}
\usepackage{stfloats}
\usepackage{url}
\usepackage{verbatim}
\usepackage{graphicx}
\usepackage{cite}
\usepackage{latexsym}
\usepackage{xspace}
\usepackage{microtype}
\usepackage[export]{adjustbox}
\usepackage{inconsolata}
\usepackage{amsmath}
\usepackage{graphicx}
\usepackage{booktabs}
\usepackage{multirow}
\usepackage{amsmath}
\usepackage{enumitem}
\usepackage{graphicx}
\usepackage{subcaption}
\usepackage[breakable]{tcolorbox}
\usepackage{fdsymbol}
\usepackage{fontawesome}
\usepackage[T1]{fontenc}    
\usepackage{url}            
\usepackage{booktabs}       
\usepackage{amsfonts}       
\usepackage{nicefrac}       
\usepackage{microtype}      
\usepackage{xcolor}         
\usepackage{algorithm}
\usepackage{algpseudocode}
\usepackage[table]{xcolor}
\usepackage{tabularx}
\newcolumntype{C}{>{\centering\arraybackslash}X}

\newtcolorbox{promptbox}[1][]{
  colback=gray!5!white,    
  colframe=gray!60!black,  
  title=#1,                
  breakable,               
  coltitle=white,          
  boxrule=1pt,           
  arc=1mm,                 
}

\usepackage{xcolor}

\usepackage{longtable}

\newcommand{\ours}{\textbf{HATS}\xspace}

\newcommand{\atree}{\textbf{a11ytree}\xspace}
\newcommand{\gpt}{\textbf{GPT-4o}\xspace}
\definecolor{veronica-red}{RGB}{196,30,58}
\definecolor{ForestGreen}{RGB}{34,139,34}
\definecolor{BrickRed}{rgb}{.72,0,0}
\definecolor{LakeBlue}{RGB}{0,61,153}

\definecolor{lightblue}{RGB}{68,14,196}
\def\BibTeX{{\rm B\kern-.05em{\sc i\kern-.025em b}\kern-.08em
    T\kern-.1667em\lower.7ex\hbox{E}\kern-.125emX}}


\usepackage{tabularx} 
\usepackage{array}    

\definecolor{cvprblue}{rgb}{0.21,0.49,0.74}
\usepackage[pagebackref,breaklinks,colorlinks,allcolors=cvprblue]{hyperref}
\def\paperID{*****} %
\def\confName{CVPR}
\def\confYear{2026}

\title{\ours: Hardness-Aware Trajectory Synthesis for GUI Agents}
\author{IEEE Publication Technology Department}

\author{
Rui Shao$^{1,3,\dagger}$, Ruize Gao$^{2,\dagger}$, Bin Xie$^{1}$, Yixing Li$^{1}$,\\
Kaiwen Zhou$^{4}$, Shuai Wang$^{4}$, Weili Guan$^{1,3}$, Gongwei Chen$^{1,*}$\\
$^{1}$Harbin Institute of Technology, Shenzhen\quad
$^{2}$National University of Singapore, CNRS@CREATE\\
$^{3}$Shenzhen Loop Area Institute\quad
$^{4}$Huawei Noah's Ark Lab\\
\texttt{shaorui@hit.edu.cn}\hspace{0.5cm}
\texttt{ruizegao@u.nus.edu}\hspace{0.5cm}
\texttt{chengongwei@hit.edu.cn}
\\
\texttt{\normalsize{\url{https://github.com/JiuTian-VL/HATS}}}
}

\begin{document}
\maketitle

\begingroup
\renewcommand\thefootnote{}
\footnotetext{$^{\dagger}$Equal contribution $^{*}$Corresponding author}
\endgroup

\begin{abstract}
Graphical user interface (GUI) agents powered by large vision–language models (VLMs) have shown remarkable potential in automating digital tasks, highlighting the need for high-quality trajectory data to support effective agent training. Yet existing trajectory synthesis pipelines often yield agents that fail to generalize beyond simple interactions. We identify this limitation as stemming from the neglect of semantic-ambiguous actions—interactions whose meanings are context-dependent, sequentially dependent, or visually ambiguous. Such actions are crucial for real-world robustness but are under-represented and poorly processed in current datasets, leading to semantic misalignment between task instructions and execution. To address these issues, we propose \textbf{\ours}, a \textbf{H}ardness-\textbf{A}ware \textbf{T}rajectory \textbf{S}ynthesis framework designed to mitigate the impact of semantic ambiguity. We define hardness as the degree of semantic ambiguity associated with an action and develop two complementary modules: (1) a hardness-driven exploration that guides data collection toward ambiguous yet informative interactions, and (2) an alignment-guided refinement that iteratively validates and repairs instruction–execution alignment. The two modules operate in a closed-loop manner—exploration supplies refinement with challenging trajectories, while refinement feedback updates the hardness signal to guide future exploration. Extensive experiments show that agents trained with \ours consistently outperform state-of-the-art baselines across benchmark GUI environments.
\end{abstract}

\begin{figure}[t]
  \centering
  \includegraphics[width=0.95\linewidth]{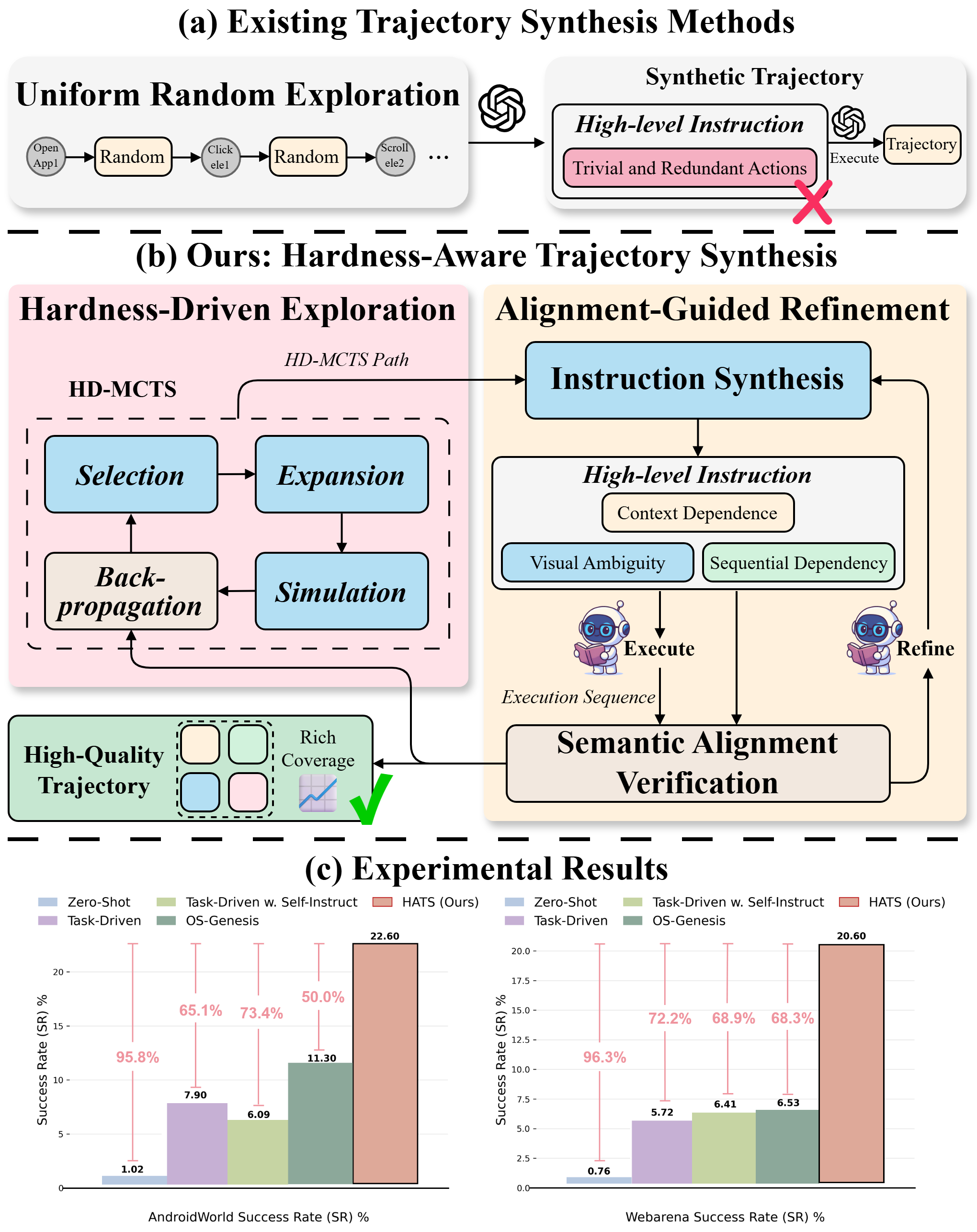}
  \caption{Overview of trajectory synthesis paradigms. Compared with (a) existing methods, (b) \ours integrates hardness-driven exploration and alignment-guided refinement in a closed loop, producing high-quality trajectories with rich semantic coverage and strong instruction--execution alignment. (c) Experiments show \ours{} outperforms \textsc{OS-Genesis} by \textbf{100\%↑} on \textsc{AndroidWorld} (\textbf{22.60 vs. 11.30}) and \textbf{215\%↑} on \textsc{WebArena} (\textbf{20.60 vs. 6.53}).}
  \label{fig:teaser_overview}
  \vspace{-1em}
\end{figure}

\section{Introduction}
\textbf{Graphical User Interface} (GUI) agents powered by large \textbf{Vision–Language Models} (VLMs) have emerged as a promising direction for automating complex digital workflows, from booking travel in mobile apps to configuring cloud services in browsers. A capable agent must perceive the screen, interpret a natural-language goal, decompose it into a plan, and execute actions such as \textsc{Tap}, \textsc{Scroll}, or \textsc{Type}. Training such autonomy requires instruction-aligned trajectories coupling task goals, step-level descriptions, and exact action–state pairs~\citep{zheng2024synapse}. However, collecting such trajectories manually is prohibitively expensive, motivating recent efforts toward automated trajectory synthesis.

\vspace{0.4em}
\noindent\textbf{Automating Trajectory Synthesis.}
To reduce human annotation, recent work automates trajectory generation. Early \textbf{task-driven} approaches begin with written goals and record agents or humans executing them~\citep{li2024androidcontrol,lu2024weblinx,lai2024autowebglm}, but still depend on costly goal specification and yield inconsistent traces~\citep{murty2024bagel,patel2024large}. \textsc{OS-Genesis}~\citep{sun2024genesis} instead introduces \textbf{reverse task synthesis}, where the system explores applications via random UI actions and then synthesizes task instructions to describe observed outcomes, improving diversity by removing manual goal authoring.

\vspace{0.4em}
\noindent\textbf{The Semantic-Ambiguity Gap.}
Despite these advantages, reverse task synthesis~\citep{sun2024genesis} still faces critical limitations. 
Exploration based on uniform random walks or shallow breadth-first search causes over 70\% of collected traces to collapse into trivial \textbf{semantic-intuitive actions} (e.g., ``open menu'', ``tap back''), which quickly saturate learning yet dominate storage and training budgets. 
Conversely, the resulting corpora remain deficient in \textbf{semantic-ambiguous actions}—UI interactions whose functional meaning depends on subtle visual or contextual cues and are easily confused or hidden.

\begin{figure}[!t]
  \centering
  \includegraphics[width=\linewidth]{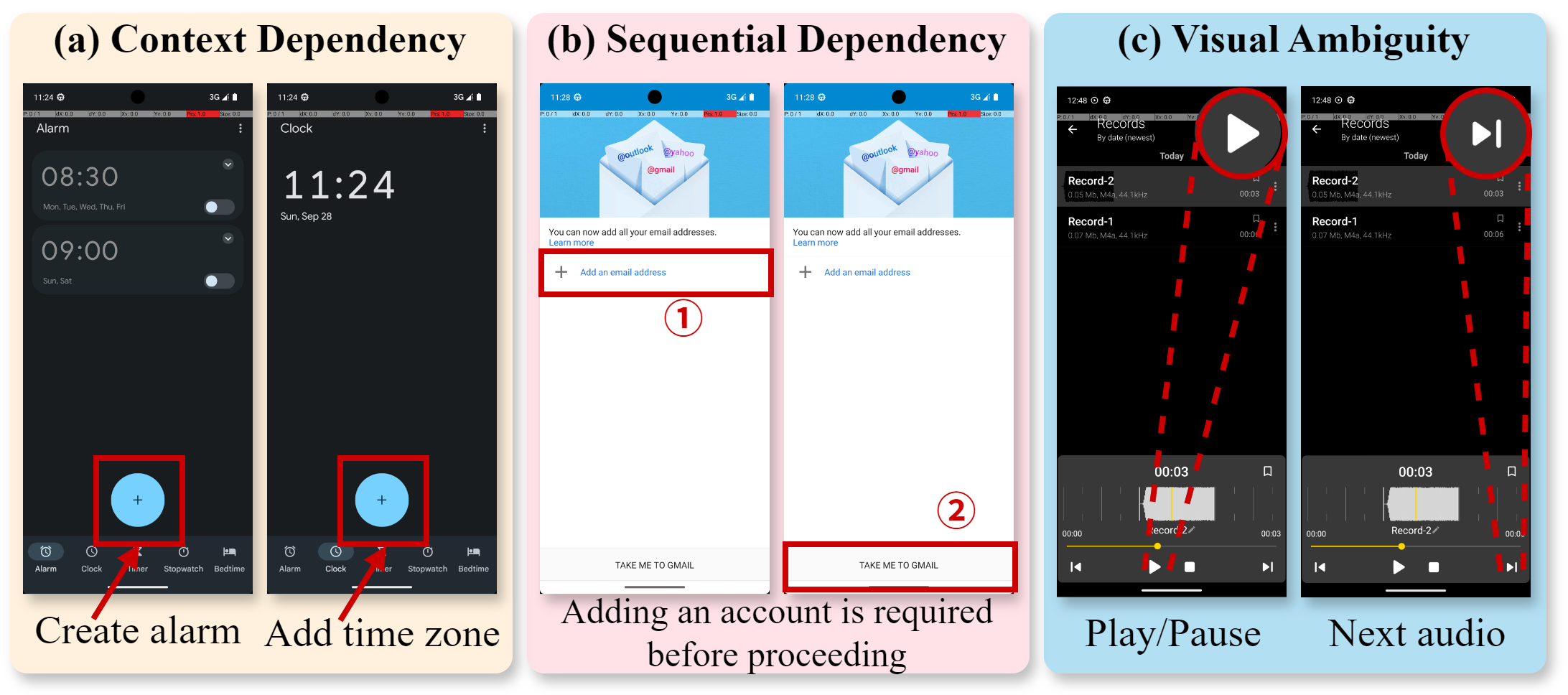}
  \caption{Illustrative cases of semantic-ambiguous actions. Such actions constitute critical bottlenecks for robust agent generalization but are rarely captured by existing synthesis pipelines.}
  \label{fig:asa_examples}
  \vspace{-1em}
\end{figure}

As illustrated in Fig.~\ref{fig:asa_examples}, these actions include (a) identical ``plus'' icons that trigger different functions depending on context, (b) operations that only succeed after prerequisite steps, and (c) visually similar elements leading to distinct outcomes. 
Such rare yet crucial actions represent the key bottlenecks where agents fail to generalize: without adequate exposure, they plateau at superficial behaviors and cannot handle unseen workflows. 
Moreover, when these actions do appear, their ambiguity often leads to \textbf{instruction--execution misalignment}: synthesized instructions may omit essential context or replay inconsistently, introducing noisy supervision that further degrades data quality. This issue is exacerbated by existing one-shot instruction generation methods~\citep{sun2024genesis}, which lack semantic validation and thus produce corpora that are both redundant and unreliable. The \textbf{challenge} is thus twofold: (i) how to focus exploration on high-value, semantically ambiguous regions of the state space rather than collecting redundant, intuitive actions; and (ii) how to ensure alignment between synthesized instructions and actual executions to prevent supervision noise. Existing one-shot instruction generation lacks semantic validation, making the resulting corpora both redundant and unreliable.

\vspace{0.4em}
\noindent\textbf{Hardness-Aware Trajectory Synthesis (\ours).}
To address these challenges, we propose \ours, a closed-loop trajectory synthesis framework explicitly targeting semantic-ambiguous actions through two cooperative modules (Fig.~\ref{fig:teaser_overview}). 
The \textbf{Hardness-Driven Exploration} module (\S\ref{sec_module_explore}) steers search toward semantically complex, under-represented states, 
while the \textbf{Alignment-Guided Refinement} module (\S\ref{sec_module_alignment}) performs multi-round replay and repair to ensure semantic consistency before data admission. 
Both modules operate under a unified \textbf{Hardness-Driven Monte Carlo Tree Search (HD-MCTS)} procedure, 
which orchestrates exploration, refinement, and feedback propagation in a single control loop.  
Exploration supplies challenging trajectories for validation, and refinement converts misalignment feedback into hardness signals that guide subsequent exploration.  
This closed-loop design systematically enhances both the diversity and fidelity of synthesized data, yielding more reliable and generalizable GUI agents.

\begin{figure*}[!t]
\vspace{-6pt}
  \centering
  \includegraphics[width=0.95\linewidth]
  {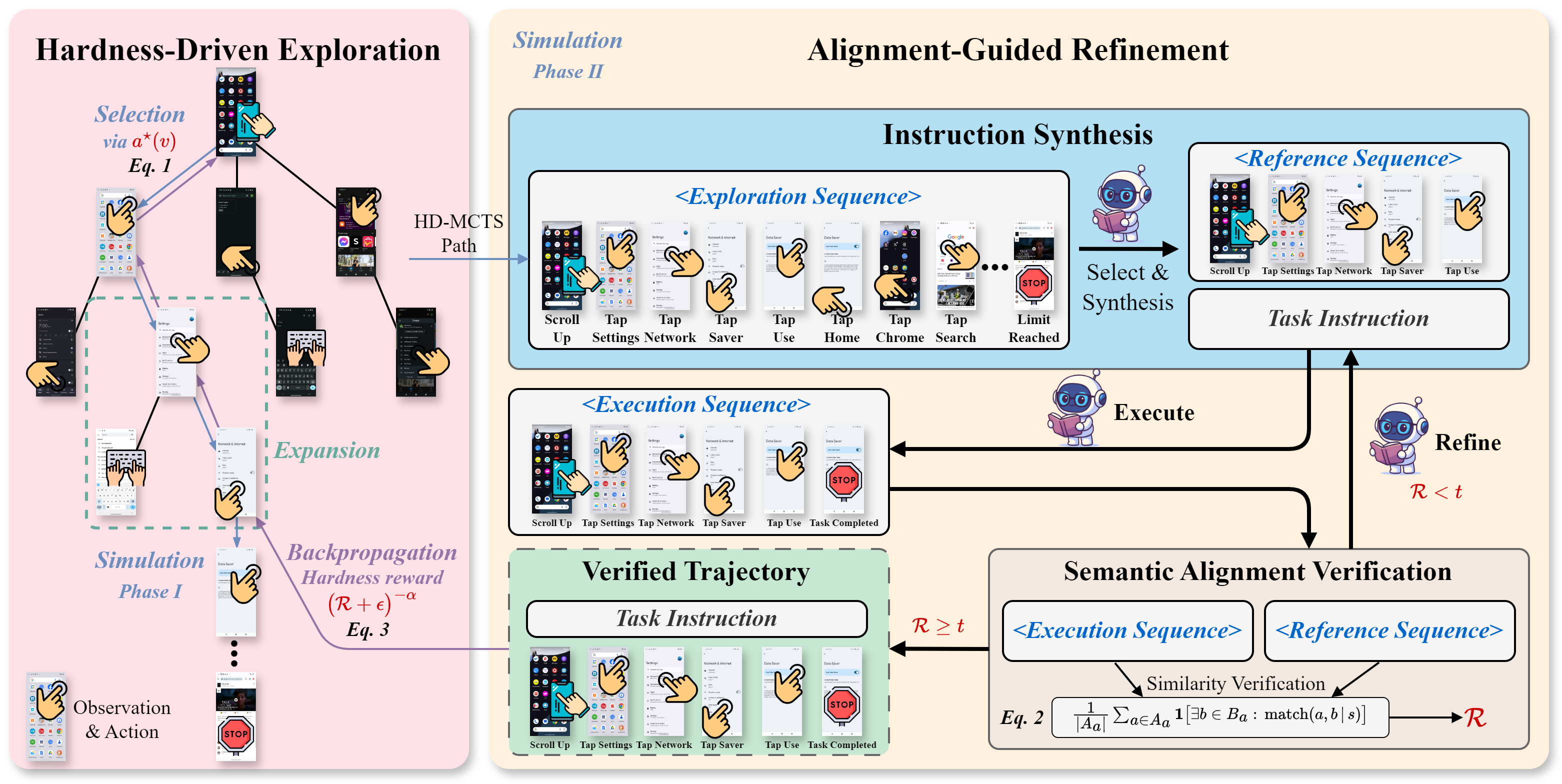}
  \caption{
    \textbf{Architecture of the \ours framework.}  
    The framework integrates a \textbf{Hardness-Driven Exploration} module (\S\ref{sec_module_explore}) and an \textbf{Alignment-Guided Refinement} module (\S\ref{sec_module_alignment}) within a unified \textbf{HD-MCTS} loop.  
    Exploration corresponds to the \textbf{Selection}, \textbf{Expansion}, and  \textbf{Simulation Phase I}, while refinement handles \textbf{Simulation Phase II} and \textbf{Backpropagation}.  
    Misalignment detected during refinement is converted into a hardness reward that guides subsequent exploration, forming a closed loop for progressively improving both diversity and semantic fidelity of synthesized trajectories.}
  \label{fig:ours_overview}
  \vspace{-1em}
\end{figure*}
\vspace{0.4em}
\noindent\textbf{Contributions.}  
We summarize our contributions as follows:
\begin{itemize}[leftmargin=1.5em]
\item \textbf{I.} We introduce \ours, a unified closed-loop trajectory synthesis framework that enhances the representation and handling of semantic-ambiguous actions.
\item \textbf{II.} We design a hardness-driven exploration module that prioritizes rare, high-value actions while reducing redundancy from intuitive ones.
\item \textbf{III.} We propose an alignment-guided refinement module that replaces one-shot synthesis with multi-round validation, ensuring instruction–execution alignment.
\end{itemize}
These contributions collectively establish a scalable, semantically grounded pipeline that enhances diversity and fidelity in trajectory synthesis, advancing GUI agent training.

\section{Related Works}
\noindent\textbf{GUI Agents.}
Recent advances in large language and vision--language models (LLMs/VLMs) have accelerated progress in agents that perceive, reason, and act within graphical user interfaces~\citep{durante2024agent,feng2024far,wu2024oscopilot}. 
Earlier systems relied on prompt programming or external tool execution~\citep{wu2023autogen,sun2024survey}, while more recent work fine-tunes multimodal backbones to better recognize UI states~\citep{cheng2024seeclick,wu2024atlas} and execute atomic actions across web, desktop, and mobile environments~\citep{kapoor2024omniact,li2024androidcontrol,wang2024mobileagent}. 
Although these agents demonstrate strong generality, their performance remains heavily constrained by the quality and diversity of available trajectory data, especially when handling semantically complex interactions.

\vspace{0.4em}
\noindent\textbf{GUI Trajectory Synthesis.}
A major challenge in data construction is obtaining large-scale, instruction-aligned trajectories that reflect realistic interface behaviors~\citep{deka2017rico,shi2017world,rawles2024androidworld,zhou2024webarena}.  
Existing approaches follow two paradigms.  
\textsc{Task-driven} demonstrations~\citep{li2024androidcontrol,lu2024weblinx} require manually authored goals and human/model executions, but are expensive and often yield inconsistent traces~\citep{murty2024bagel,patel2024large}.  
To improve scalability, \textsc{OS-Genesis}~\citep{sun2024genesis} adopts an \textbf{exploration-driven} strategy, exploring UIs first and then synthesizing task instructions from observed outcomes.  
However, naive exploration oversamples repetitive \textbf{semantic-intuitive actions} while underrepresenting \textbf{semantic-ambiguous actions} whose functional meaning depends on contextual, sequential, or visual cues.  
Moreover, one-shot reverse synthesis frequently produces underspecified instructions that fail to replay, leading to \textbf{instruction-execution misalignment}. These limitations motivate our hardness-aware, closed-loop framework, which targets ambiguous interactions during exploration and verifies synthesized instructions through iterative alignment. Other contemporaneous efforts explore stochastic or intent-aware multi-app exploration~\citep{lin2025gui}, which is largely orthogonal to our focus on addressing semantic ambiguity through hardness-aware sampling and alignment refinement.

\section{Hardness-Aware Trajectory Synthesis}
\label{sec:method}
Existing automated pipelines struggle in two key ways: (i) they rarely capture semantic-ambiguous actions, i.e., interactions whose semantics depend on contextual, sequential, or visual cues; and (ii) even when such actions are captured, one-shot instruction generation often produces noisy or misaligned descriptions that degrade supervision quality.  
Our framework \ours addresses both issues by tightly coupling targeted exploration with alignment-aware data verification.

\vspace{0.4em}
\noindent\textbf{Method Overview.}
As illustrated in Fig.~\ref{fig:ours_overview}, \ours consists of two interacting modules:  
a \textbf{Hardness-Driven Exploration} module that actively seeks semantically ambiguous, high-value actions (\S\ref{sec_module_explore}), and  
an \textbf{Alignment-Guided Refinement} module that validates and repairs the resulting instruction–trajectory pairs before admitting them into the training corpus (\S\ref{sec_module_alignment}).  
These two modules are fused through a customized \textbf{Hardness-Driven Monte Carlo Tree Search} (HD-MCTS) procedure that forms the unified control loop of our framework. Instead of treating exploration and supervision cleanup as two separate offline stages,  
\ours unifies them into a single iterative search process.  
Each HD-MCTS iteration (Appendix Alg.~\ref{alg:hdmcts_final}) performs four coordinated stages:
\begin{itemize}[leftmargin=1.5em]
\item \textbf{Selection}: the agent traverses an action tree to balance exploitation and exploration using a UCB policy;
\item \textbf{Expansion}: the tree grows by executing previously unvisited actions and adding corresponding nodes;
\item \textbf{Simulation}: the agent rolls out a bounded-depth trajectory, reverse-synthesizes a natural-language instruction, replays and \textbf{iteratively refines} that instruction until it becomes executable and semantically aligned;
\item \textbf{Backpropagation}: an \textbf{alignment-derived hardness reward} is propagated up the tree to bias future exploration toward semantically challenging regions.
\end{itemize}

\vspace{0.4em}
\noindent\textbf{Notations.}  
Let $\mathcal{E}$ denote a GUI environment.  
At time $t$, the agent observes a state $s_t$ (screenshot + optional UI tree), executes an action $a_t \in \mathcal{A}$, and transitions to $s_{t+1} = \mathcal{E}(s_t, a_t)$.  
The atomic action set is  
$\mathcal{A} = \{\textsc{Tap}, \textsc{Type}, \textsc{Scroll}, \textsc{Back}, \textsc{Long\mbox{-}Press}, \textsc{Swipe}\}$.  
A trajectory $\tau = (s_0, a_0, s_1, \dots, s_T)$ records a sequence of interactions.  
We represent $\mathcal{E}$ as an \textbf{action tree}, where each node $v$ corresponds to a state $s(v)$ and each edge $(v,a)$ maintains a value estimate $Q(v,a)$ and visit count $N(v,a)$, with $N(v)=\sum_a N(v,a)$.  
$\mathcal{P}$ denotes the current path (a sequence of $(a,s)$ pairs) with rollout depth limit $T_{\max}$.
\begin{figure}[!t]
  \centering
  \includegraphics[width=\linewidth]{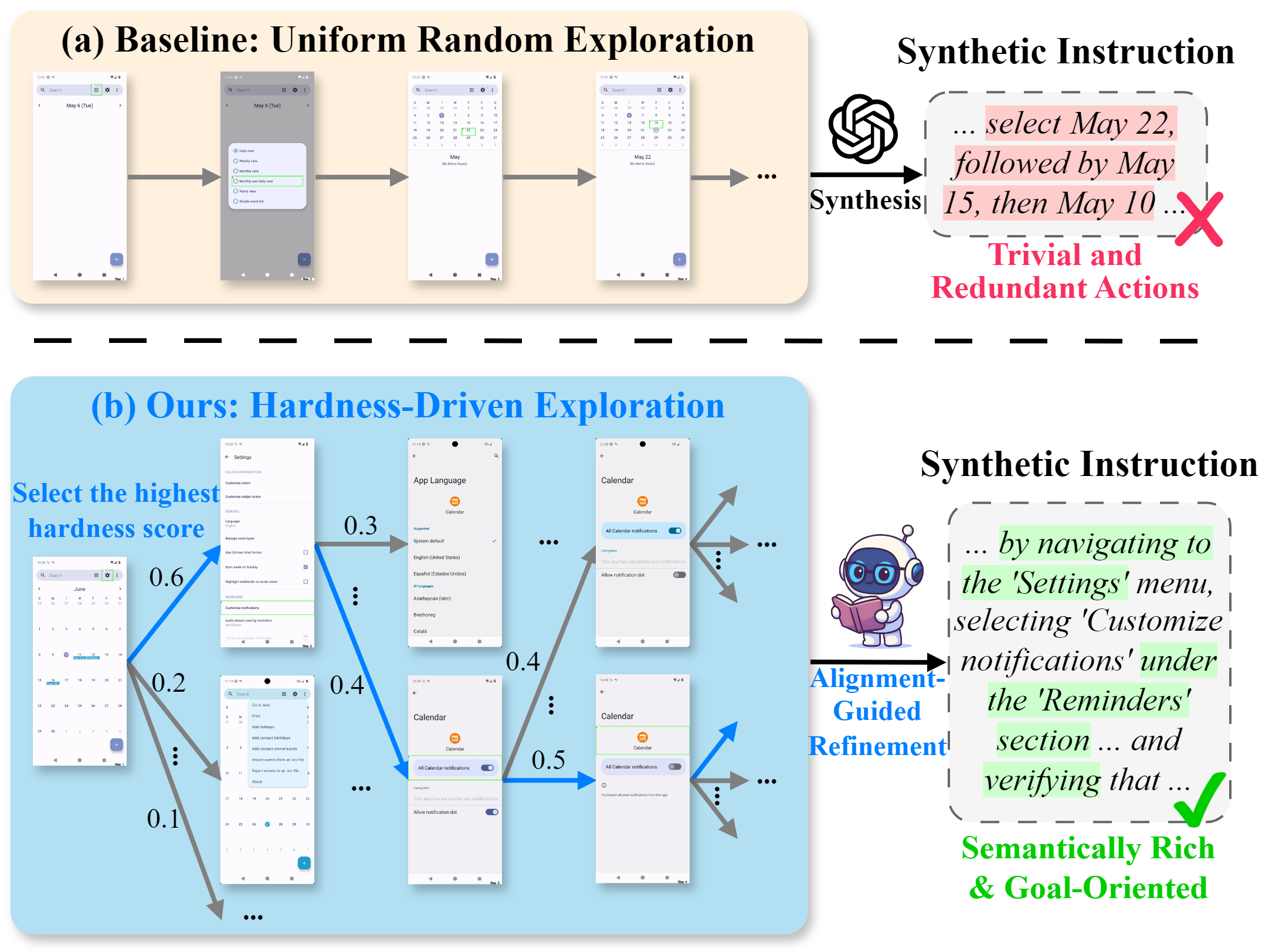}
  \caption{
    Comparison of exploration strategies. 
    \textbf{Uniform Random Exploration} \citep{sun2024genesis} often yields trivial and redundant actions, whereas our  \textbf{Hardness-Driven Exploration} replaces random walking with a hardness-driven exploration policy that selectively targets under-represented yet semantically challenging actions.}
  \label{fig:exploration_comparison}
  \vspace{-1em}
\end{figure}
\subsection{Hardness-Driven Exploration Module}
\label{sec_module_explore}
We follow the \textbf{interaction-driven discovery paradigm} introduced by \textsc{OS-Genesis}~\citep{sun2024genesis}, 
where the agent explores by performing UI actions rather than following pre-defined goals. As contrasted in Fig.~\ref{fig:exploration_comparison}, however, \textsc{OS-Genesis} performs \textbf{reverse task synthesis} from randomly sampled traces, 
often yielding trivial and redundant actions.  
In contrast, our approach replaces random walking with a \textbf{hardness-driven exploration policy} that selectively targets under-represented yet semantically challenging actions.  
This module corresponds to the \textbf{Selection} and \textbf{Expansion} phases, as well as \textbf{Simulation Phase I} in HD-MCTS.

\vspace{0.4em}
\noindent\textbf{I. Selection on the Action Tree.}
Starting from the root node $v_0$ with the initial GUI state $s_0$, 
HD-MCTS traverses the action tree by recursively selecting outgoing edges using an Upper Confidence Bound (UCB) rule:
\begin{equation}
\label{eq:ucb}
a^\star(v) \;=\; \arg\max_{a\in\mathcal{A}(v)} 
Q(v,a)
+\;
C\,\sqrt{\tfrac{\ln(N(v)+1)}{N(v,a)+1}},
\end{equation}
where $C$ controls exploration pressure. Unlike classical MCTS, $Q(v,a)$ is updated via the \textbf{hardness reward} (Eq.~\ref{eq:hardness}), which we formally define later in the closed-loop integration stage.  
This reward serves as an innovative bridge between exploration and semantic feedback—transforming alignment hardness into a guiding signal that steers future search toward more informative, ambiguous behaviors.

\vspace{0.4em}
\noindent\textbf{II. Expansion.}
If the current node $v$ is not fully expanded, we select an unvisited action $a_e \in \mathcal{A}(v)$, 
apply it to obtain $s_e = \mathcal{E}(s(v), a_e)$, create a child node $u$ with state $s(u)=s_e$, 
and append $(a_e,s_e)$ to $\mathcal{P}$.  
This grows the search frontier along under-explored branches and increases coverage of semantically diverse UI states.

\vspace{0.4em}
\noindent\textbf{III. Simulation Phase I: Rollout.}
After selection and expansion, we perform a bounded-depth \textbf{rollout} from the current node by executing valid UI actions until $\mathcal{P}$ reaches depth $T_{\max}$:
\[
\mathcal{P} \leftarrow \textsc{Rollout}(\mathcal{E}, v, s, \mathcal{P}, T_{\max}).
\]
This process produces a concrete multi-step interaction trace, a candidate trajectory, which is then passed to the \textbf{alignment-guided refinement module} for subsequent processing.

\begin{figure}[!t]
  \centering
  \includegraphics[width=\linewidth]{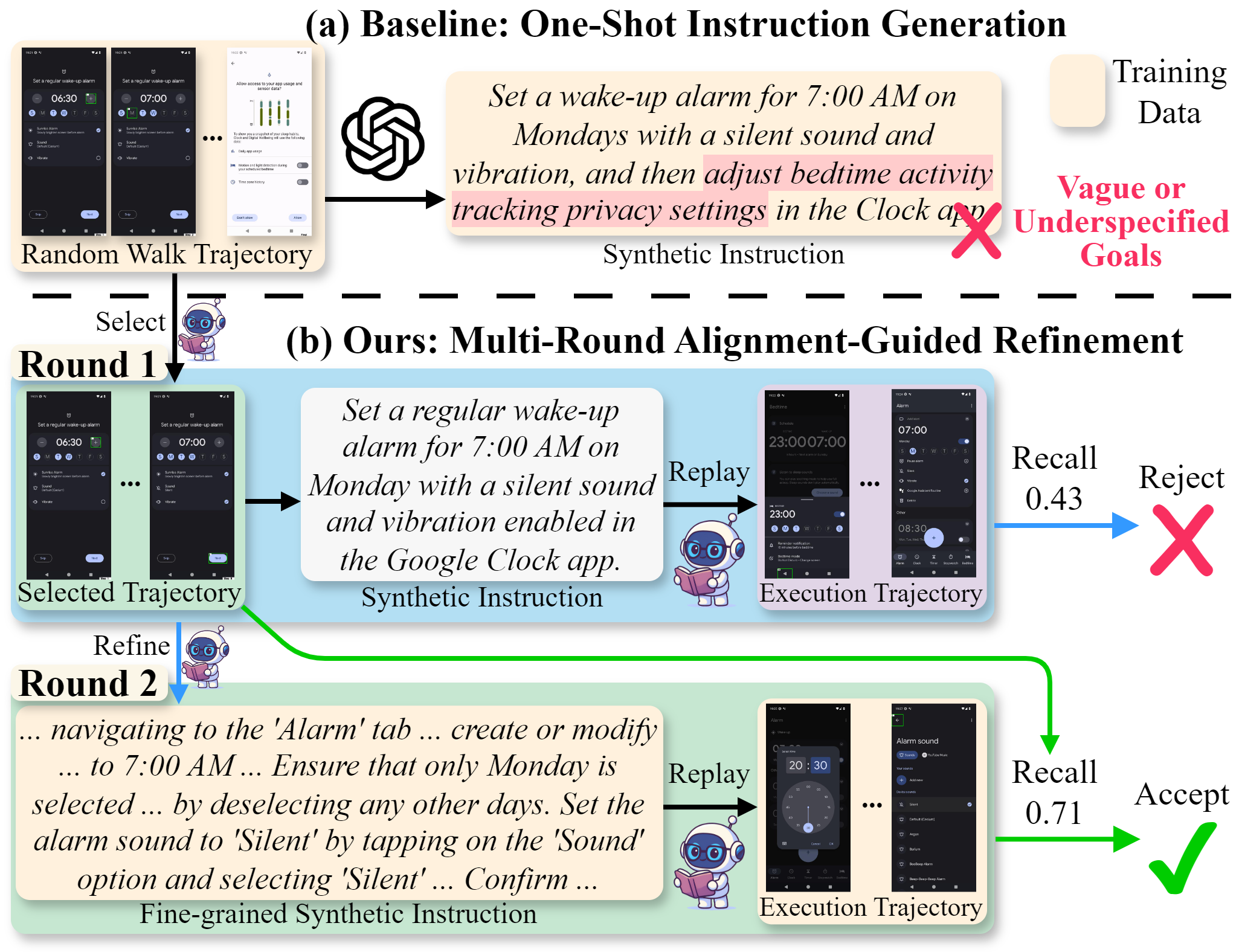}
  \caption{Comparison of instruction synthesis methods. \textbf{One-Shot Instruction Generation} \citep{sun2024genesis} directly maps raw traces to text, often yielding vague or underspecified goals and inconsistent executions. In contrast, our \textbf{Multi-Round Alignment-Guided Refinement} iteratively replays and verifies task instructions to produce semantically faithful and executable Verified Trajectories.}
  \label{fig:refinement_comparison}
  \vspace{-1em}
\end{figure}

\subsection{Alignment-Guided Refinement Module}
\label{sec_module_alignment}
This module corresponds to \textbf{Simulation Phase II} in HD-MCTS 
(see Fig.~\ref{fig:ours_overview}).  
It receives noisy, partially meaningful \textbf{Exploration Sequences} from the previous stage 
and transforms them into executable, instruction-aligned
\textbf{Verified Trajectory}.  
Unlike the \textbf{One-Shot Instruction Generation} paradigm in \textsc{OS-Genesis}~\citep{sun2024genesis}, 
which directly maps random traces to text, 
our method employs a \textbf{Multi-Round Alignment-Guided Refinement} process 
that iteratively replays, verifies, and corrects synthesized instructions 
until semantic and functional consistency are both achieved (Fig.~\ref{fig:refinement_comparison}).

\vspace{0.4em}
\noindent\textbf{{Prompt-Based Implementation.}}
All stages of refinement that sub-trajectory selection, instruction synthesis,
execution replay, similarity verification, and contextual correction are
implemented via prompt-driven VLM modules \footnotemark[1].

\vspace{0.4em}
\noindent\textbf{I. Sub-Trajectory Selection.}  
From the rolled-out \textbf{Exploration Sequence} $\mathcal{P}$,  
we first extract a semantically coherent segment:
\begingroup
\setlength{\abovedisplayskip}{4pt} 
\setlength{\belowdisplayskip}{4pt} 
\[
A \gets \textsc{SelectSubsequence}(\mathcal{P})\footnotemark[1],
\]
\endgroup
representing one meaningful user intent, such as ``add a new clock''.  
This selected segment is referred to as the \textbf{Reference Sequence}, 
serves as the behavioral ground truth for synthesis and verification.

\vspace{0.4em}
\noindent\textbf{II. Instruction Synthesis.}  
We then generate a natural-language description for the Reference Sequence via
\begingroup
\setlength{\abovedisplayskip}{4pt} 
\setlength{\belowdisplayskip}{4pt} 
\[
I \gets \textsc{SynthesizeInstruction}(A)\footnotemark[1],
\]
\endgroup
yielding an initial hypothesized instruction.  
Unlike \textsc{OS-Genesis}, which treats this output as final,   
we regard $I$ as provisional and subject to replay-driven verification.

\begin{table*}[t]
\centering
\scriptsize
\setlength{\tabcolsep}{4pt}
\caption{
\textbf{AndroidWorld evaluation results across task categories. All values are success rates (\%). \textbf{DLS} = Daily Life \& Services; \textbf{S\&C} = Social \& Communication; \textbf{S\&U} = System \& Utility; \textbf{P\&W} = Productivity \& Work. The final column shows overall success rate.}}
\vspace{-8pt}
\begin{tabularx}{\linewidth}{ll|CCCC|C} 
\toprule[1.5pt]

\multirow{2}{*}{\textbf{Base Model}} & \multirow{2}{*}{\textbf{Strategies}} & \multicolumn{4}{c|}{\textbf{Task Categories (\%)}} & \multirow{2}{*}{\textbf{Overall SR (\%)}} \\
 &  & \textbf{DLS} & \textbf{S\&C} & \textbf{S\&U} & \textbf{P\&W} & \\
\midrule[0.1pt]
\midrule[0.1pt]
\rowcolor[HTML]{FFF9E6} \textbf{GPT-4o~\citep{hurst2024gpt}} & Zero-Shot~\citep{wei2022chain} & 42.86\,{\tiny N.A.} & 66.67\,{\tiny N.A.} & 77.27\,{\tiny N.A.} & 29.09\,{\tiny N.A.} & 45.22\,{\tiny N.A.} \\
\midrule
\multirow{5}{*}{InternVL2-4B~\citep{chen2023internvl}} 
 & Zero-Shot~\citep{wei2022chain} & 0.00\,{\tiny -100.0\%} & 0.00\,{\tiny -100.0\%} & 2.38\,{\tiny -94.2\%} & 0.00\,{\tiny -100.0\%} & 1.02\,{\tiny -95.8\%} \\
 & Task-Driven~\citep{lai2024autowebglm} & 3.49\,{\tiny -79.8\%} & 0.00\,{\tiny -100.0\%} & 19.70\,{\tiny -51.8\%} & 3.70\,{\tiny -57.2\%} & 7.90\,{\tiny -65.1\%} \\
 & Task-Driven w. Self-Instruct~\citep{wang2023selfinstruct} & 2.30\,{\tiny -86.7\%} & 3.70\,{\tiny -90.4\%} & 21.21\,{\tiny -48.2\%} & 0.62\,{\tiny -92.8\%} & 6.09\,{\tiny -73.4\%} \\
 & OS-Genesis~\citep{sun2024genesis} & 7.14\,{\tiny -58.6\%} & 11.11\,{\tiny -71.2\%} & 36.36\,{\tiny -11.1\%} & 3.64\,{\tiny -57.9\%} & 11.30\,{\tiny -50.0\%} \\
\cmidrule(lr){2-7}
\rowcolor[HTML]{E6F3FF}  & \ours & \textbf{17.24}\,{\tiny 0.0\%} & \textbf{38.46}\,{\tiny 0.0\%} & \textbf{40.91}\,{\tiny 0.0\%} & \textbf{8.64}\,{\tiny 0.0\%} & \textbf{22.60}\,{\tiny 0.0\%}\\
\midrule
\multirow{5}{*}{InternVL2-8B~\citep{chen2023internvl}} 
 & Zero-Shot~\citep{wei2022chain} & 0.00\,{\tiny -100.0\%} & 0.00\,{\tiny -100.0\%} & 3.13\,{\tiny -94.1\%} & 0.00\,{\tiny -100.0\%} & 1.04\,{\tiny -95.8\%} \\
 & Task-Driven~\citep{lai2024autowebglm} & 1.15\,{\tiny -92.9\%} & 0.00\,{\tiny -100.0\%} & 12.12\,{\tiny -77.2\%} & 0.62\,{\tiny -95.4\%} & 3.48\,{\tiny -85.7\%} \\
 & Task-Driven w. Self-Instruct~\citep{wang2023selfinstruct} & 5.95\,{\tiny -63.1\%} & 0.00\,{\tiny -100.0\%} & 13.64\,{\tiny -74.3\%} & 3.09\,{\tiny -77.2\%} & 6.25\,{\tiny -72.4\%} \\
 & OS-Genesis~\citep{sun2024genesis} & 13.79\,{\tiny -14.3\%} & 0.00\,{\tiny -100.0\%} & 47.62\,{\tiny -10.2\%} & 4.08\,{\tiny -70.0\%} & 14.81\,{\tiny -39.4\%} \\
\cmidrule(lr){2-7}
\rowcolor[HTML]{E6F3FF}  & \ours & \textbf{16.09}\,{\tiny 0.0\%} & \textbf{25.93}\,{\tiny 0.0\%} & \textbf{53.03}\,{\tiny 0.0\%} & \textbf{13.58}\,{\tiny 0.0\%} & \textbf{24.35}\,{\tiny 0.0\%} \\
\midrule
\multirow{5}{*}{Qwen2-VL-7B~\citep{Qwen2VL}} 
 & Zero-Shot~\citep{wei2022chain} & 0.00\,{\tiny -100.0\%} & 0.00\,{\tiny -100.0\%} & 7.58\,{\tiny -85.3\%} & 0.00\,{\tiny -100.0\%} & 1.75\,{\tiny -92.6\%} \\
 & Task-Driven~\citep{lai2024autowebglm} & 8.33\,{\tiny -54.7\%} & 3.70\,{\tiny -88.0\%} & 30.30\,{\tiny -41.2\%} & 2.47\,{\tiny -75.1\%} & 10.43\,{\tiny -57.2\%} \\
 & Task-Driven w. Self-Instruct~\citep{wang2023selfinstruct} & 6.90\,{\tiny -62.5\%} & 7.41\,{\tiny -75.9\%} & 25.76\,{\tiny -50.0\%} & 0.62\,{\tiny -93.7\%} & 6.09\,{\tiny -74.9\%} \\
 & OS-Genesis~\citep{sun2024genesis} & 17.86\,{\tiny -3.0\%} & 0.00\,{\tiny -100.0\%} & 31.82\,{\tiny -38.2\%} & 1.85\,{\tiny -81.3\%} & 11.50\,{\tiny -52.7\%} \\
\cmidrule(lr){2-7}
\rowcolor[HTML]{E6F3FF}  & \ours & \textbf{18.39}\,{\tiny 0.0\%} & \textbf{30.77}\,{\tiny 0.0\%} & \textbf{51.52}\,{\tiny 0.0\%} & \textbf{9.88}\,{\tiny 0.0\%} & \textbf{24.35}\,{\tiny 0.0\%} \\
\bottomrule[1.5pt]
\end{tabularx}
\label{tab:android_results}
\vspace{-1em}
\end{table*}
\vspace{0.4em}
\noindent\textbf{III. Replay and Execution.}  
We execute the synthesized instruction $I$ from the start state of $A$ to obtain
an \textbf{Execution Sequence}:
\begingroup
\setlength{\abovedisplayskip}{4pt} 
\setlength{\belowdisplayskip}{4pt} 
\[
B \gets \textsc{ExecuteInstruction}(I)\footnotemark[1].
\]
\endgroup
This sequence represents how the environment actually responds to the instruction.  
Discrepancies between $A$ (expected behavior) and $B$ (observed outcome) 
indicate semantic gaps in $I$, e.g., missing contextual cues or ambiguity.

\vspace{0.4em}
\noindent\textbf{IV. Similarity Verification.}  
To quantitatively measure alignment between the \textbf{Reference} and \textbf{Execution Sequences},  
we perform the similarity verification using the \textbf{Action-Level Reconstruction Recall} metric:
\begin{equation}
\label{eq:recall}
R(A,B)
=\frac{1}{|A_a|}
\sum_{a\in A_a}
\mathbf{1}\bigl[\exists b\in B_a:\,\mathrm{match}(a,b\mid s)\bigr].
\end{equation}
Here, $\mathrm{match}(a,b\mid s)$\footnotemark[1] determines whether executing $b$ in state $s$  
is semantically equivalent to action $a$ in its original context.  
A lower $R(A,B)$ implies that the instruction $I$ lacks critical contextual cues,
for instance, it may omit widget identifiers, temporal dependencies, or navigation context.
We later compare this recall metric with a precision-based alternative in our 
metric analysis (\S\ref{sec:exp_analysis}).

\footnotetext[1]{The prompt templates used at each stage are provided in Appendix~\ref{app:prompt_summary}.}

\vspace{0.4em}
\noindent\textbf{V. Iterative Refinement and Final Verification.} If $R(A,B) < R_{\min}$ or execution fails, $I$ is refined by injecting missing context and re-executed to obtain an updated \textbf{Execution Sequence}.\footnotemark[1] This replay–refine loop continues until alignment improves ($R(A,B) \ge R_{\min}$) or the refinement budget $F_{\max}$ is reached. We set $R_{\min}=0.7$ based on empirical validation balancing stability and coverage. When $I$ yields stable, deterministic execution, the pair $(I,B)$ is accepted as a \textbf{Verified Trajectory}. Otherwise, it is discarded, and its misalignment contributes to the \textbf{hardness reward} for future exploration.

Only verified trajectories that pass alignment verification are admitted to the corpus, ensuring that synthesized data are both \textbf{reliable} (free from misaligned supervision) and \textbf{informative} (rich in semantic-ambiguous yet executable behaviors).  
The effectiveness of the refinement module is analyzed in \S\ref{sec:refine_effect}, where multi-round replay is shown to monotonically increase action-level recall and substantially improve downstream task success. 
As visualized in Fig.~\ref{fig:refinement_comparison}, iterative refinement transforms vague one-shot instructions into precise and executable task instructions, bridging the gap between textual goals and GUI-level actions.

\begin{table*}[t]
\centering
\scriptsize
\caption{\textbf{WebArena evaluation results across website domains. All values are success rates (\%).}}
\vspace{-8pt}
\begin{tabularx}{\linewidth}{ll|CCCC} 
\toprule[1.5pt]
\multirow{2}{*}{\textbf{Base Model}} & \multirow{2}{*}{\textbf{Strategies}} & \multicolumn{4}{c}{\textbf{WebArena (Success Rate \%)}} \\
 &  & \textbf{Gitlab} & \textbf{Maps} & \textbf{Reddit} & \textbf{Overall} \\
\midrule[0.1pt]
\midrule[0.1pt]
\rowcolor[HTML]{FFF9E6} \textbf{GPT-4o~\citep{hurst2024gpt}} & Zero-Shot~\citep{wei2022chain} & 32.14\,{\tiny N.A.} & 6.42\,{\tiny N.A.} & 27.36\,{\tiny N.A.} & 22.82\,{\tiny N.A.} \\
\midrule
\multirow{5}{*}{InternVL2-4B~\citep{chen2023internvl}} 
 & Zero-Shot~\citep{wei2022chain} & 1.11\,{\tiny -95.1\%} & 0.00\,{\tiny -100.0\%} & 0.94\,{\tiny -94.8\%} & 0.76\,{\tiny -96.3\%} \\
 & Task-Driven~\citep{lai2024autowebglm} & 9.52\,{\tiny -58.1\%} & 5.00\,{\tiny -75.2\%} & 0.00\,{\tiny -100.0\%} & 5.72\,{\tiny -72.2\%} \\
 & Task-Driven w. Self-Instruct~\citep{wang2023selfinstruct} & 9.52\,{\tiny -58.1\%} & 7.50\,{\tiny -67.0\%} & 0.00\,{\tiny -100.0\%} & 6.41\,{\tiny -68.9\%} \\
 & OS-Genesis~\citep{sun2024genesis} & 7.94\,{\tiny -65.1\%} & 7.50\,{\tiny -67.0\%} & 3.13\,{\tiny -82.6\%} & 6.53\,{\tiny -68.3\%} \\
\cmidrule(lr){2-6}
\rowcolor[HTML]{E6F3FF}  & \ours & \textbf{22.73}\,{\tiny 0.0\%} & \textbf{20.19}\,{\tiny 0.0\%} & \textbf{17.92}\,{\tiny 0.0\%} & \textbf{20.60}\,{\tiny 0.0\%} \\
\midrule
\multirow{5}{*}{InternVL2-8B~\citep{chen2023internvl}} 
 & Zero-Shot~\citep{wei2022chain} & 1.37\,{\tiny -95.1\%} & 0.95\,{\tiny -96.2\%} & 0.97\,{\tiny -95.2\%} & 1.13\,{\tiny -95.5\%} \\
 & Task-Driven~\citep{lai2024autowebglm} & 6.35\,{\tiny -77.2\%} & 2.50\,{\tiny -89.9\%} & 0.00\,{\tiny -100.0\%} & 3.58\,{\tiny -85.6\%} \\
 & Task-Driven w. Self-Instruct~\citep{wang2023selfinstruct} & 7.94\,{\tiny -71.4\%} & 0.00\,{\tiny -100.0\%} & 6.25\,{\tiny -69.0\%} & 5.30\,{\tiny -78.7\%} \\
 & OS-Genesis~\citep{sun2024genesis} & 6.35\,{\tiny -77.2\%} & 10.00\,{\tiny -59.6\%} & 9.34\,{\tiny -53.7\%} & 8.16\,{\tiny -67.2\%} \\
\cmidrule(lr){2-6}
\rowcolor[HTML]{E6F3FF}  & \ours & \textbf{27.81}\,{\tiny 0.0\%} & \textbf{24.77}\,{\tiny 0.0\%} & \textbf{20.19}\,{\tiny 0.0\%} & \textbf{24.87}\,{\tiny 0.0\%} \\
\midrule
\multirow{5}{*}{Qwen2-VL-7B~\citep{Qwen2VL}} 
 & Zero-Shot~\citep{wei2022chain} & 1.14\,{\tiny -95.4\%} & 0.98\,{\tiny -95.4\%} & 1.96\,{\tiny -91.4\%} & 1.32\,{\tiny -94.3\%} \\
 & Task-Driven~\citep{lai2024autowebglm} & 6.35\,{\tiny -74.6\%} & 5.00\,{\tiny -76.3\%} & 6.25\,{\tiny -72.6\%} & 5.95\,{\tiny -74.4\%} \\
 & Task-Driven w. Self-Instruct~\citep{wang2023selfinstruct} & 4.84\,{\tiny -80.6\%} & 7.50\,{\tiny -64.5\%} & 3.13\,{\tiny -86.3\%} & 5.12\,{\tiny -78.0\%} \\
 & OS-Genesis~\citep{sun2024genesis} & 15.87\,{\tiny -36.5\%} & 5.00\,{\tiny -76.4\%} & 15.63\,{\tiny -31.3\%} & 12.81\,{\tiny -44.9\%} \\
\cmidrule(lr){2-6}
\rowcolor[HTML]{E6F3FF}  & \ours & \textbf{25.00}\,{\tiny 0.0\%} & \textbf{21.10}\,{\tiny 0.0\%} & \textbf{22.77}\,{\tiny 0.0\%} & \textbf{23.28}\,{\tiny 0.0\%} \\
\bottomrule[1.5pt]
\end{tabularx}
\label{tab:web_tasks}
\vspace{-1em}
\end{table*}
\subsection{Closed-Loop Integration}
\label{subsec:closed-loop}

This stage corresponds to the \textbf{Backpropagation} phase of HD-MCTS, 
where the alignment scores of refined trajectories are converted into a
unified hardness feedback signal that guides subsequent exploration.
In contrast to conventional MCTS, which backpropagates environment-provided task rewards, GUI trajectory synthesis exposes no such scalar supervision. 
We therefore propagate an \textbf{alignment-derived hardness reward} that quantifies semantic difficulty and serves as a task-adaptive surrogate objective.

\vspace{0.2em}
\noindent\textbf{I. Motivation and Design of the Hardness Reward.} The core idea is to turn semantic misalignment, typically noisy supervision, into informative feedback that prioritizes ambiguous or hard regions of the GUI state space. During refinement, we compute the \textbf{action-level reconstruction recall} $R(A,B)$ (Eq.~\ref{eq:recall}), which measures how well a synthesized instruction $I$ reproduces its reference sequence $A$. High recall indicates clear semantic grounding, whereas low recall reflects ambiguous visual cues, missing preconditions, or unstable language–interface mappings. Thus, $R$ naturally serves as an empirical estimate of \textbf{task hardness}.

Building on this insight, we define the \textbf{hardness reward} as an inverse function of $R(A,B)$:
\begin{equation}
\label{eq:hardness}
r(A,B) = (R(A,B)+\epsilon)^{-\alpha}, \quad \epsilon > 0,\ \alpha > 0.
\end{equation}
This formulation (i) yields a continuous, bounded score that increases smoothly as alignment deteriorates, and
(ii) directly links exploration to the weakest regions of instruction–execution understanding.
We additionally clip hardness into a stable range (e.g., $[0,H_{\max}]$) to avoid extreme values.
Within HD-MCTS, this reward replaces task-return signals and encourages revisiting trajectories that are hard to align yet semantically valuable. 
We use the default configuration $(\epsilon{=}0.01,\ \alpha{=}1)$, and analyze parameter sensitivity in \S\ref{sec:exp_analysis}.

\vspace{0.2em}
\noindent\textbf{II. Hardness Propagation.}
After refinement, each verified instruction–trajectory pair $(I,B)$ yields a hardness value $r(A,B)$.
We backpropagate this reward along the \textbf{tree-policy path} from the expansion node to the root:
\begingroup
\setlength{\abovedisplayskip}{6pt} 
\setlength{\belowdisplayskip}{6pt} 
\begin{equation}
N(v,a) \!\leftarrow\! N(v,a)+1,\
Q(v,a) \!\leftarrow\! Q(v,a)+\tfrac{r - Q(v,a)}{N(v,a)}.
\end{equation}
\endgroup
Following standard MCTS, only nodes selected by the tree policy receive credit.
Since hardness characterizes the entire instruction–trajectory pair rather than any single action,
the same scalar reward is propagated along the path. These updates reshape the value estimates,
biasing future UCB selection (Eq.~\ref{eq:ucb}) toward semantically challenging regions.

In summary, the closed loop of \ours turns semantic ambiguity from a training liability into a useful learning signal. Misalignment during replay generates the hardness reward, which in turn guides exploration toward richer and harder-to-align interactions. \textbf{Verified trajectories} annotated with hardness form a semantically grounded dataset for downstream SFT training. We follow the two-stage paradigm of \textsc{OS-Genesis}~\citep{sun2024genesis} for planning-level and action-level learning, with hardness-aware weighting detailed in Appendix~\ref{app:exp_details}.

\begin{table}[t]
\centering
\scriptsize
\caption{\textbf{Parameter sensitivity of the hardness reward in Eq.~\ref{eq:hardness}.}}
\vspace{-8pt}
\renewcommand{\arraystretch}{1}
\setlength{\tabcolsep}{5pt}
\label{tab:metric_param_ablation}
\begin{tabularx}{\linewidth}{
>{\centering\arraybackslash}p{0.27\linewidth}|
>{\centering\arraybackslash}p{0.13\linewidth}
>{\centering\arraybackslash}p{0.13\linewidth}|
>{\centering\arraybackslash}p{0.295\linewidth}}
\toprule[1.5pt]
\multirow{2}{*}{\textbf{Hardness Metric}} &
\multicolumn{2}{c|}{\textbf{Hyperparameters}} &
\multirow{2}{*}{\textbf{Success Rate (\%)}} \\
 & $\boldsymbol{\epsilon}$ & $\boldsymbol{\alpha}$ &  \\[2pt]
\midrule[0.1pt]
\midrule[0.1pt]
\multirow{6}{*}{$(R(A,B)+\epsilon)^{-\alpha}$}
& 0.01 & 0.50 & 22.7\,{\tiny -55.3\%} \\
\rowcolor[HTML]{E6F3FF}
& 0.01 & 1.00 & \textbf{50.8}\,{\tiny 0.0\%} \\
& 0.01 & 2.00 & 39.2\,{\tiny -22.8\%} \\
& 0.10 & 0.50 & 13.8\,{\tiny -72.8\%} \\
& 0.10 & 1.00 & 40.0\,{\tiny -21.3\%} \\
& 0.10 & 2.00 & 37.9\,{\tiny -25.3\%} \\
\bottomrule[1.5pt]
\end{tabularx}
\vspace{-1em}
\end{table}
\section{Experiments}
\label{sec:exp_analysis}

We evaluate \ours{} on two representative GUI benchmarks:
\textsc{AndroidWorld}~\citep{rawles2024androidworld} and 
\textsc{WebArena}~\citep{zhou2024webarena}, 
which together span diverse mobile and web interaction patterns. 
On \textsc{AndroidWorld}, all 116 tasks are grouped into four functional categories, 
\textbf{DLS}, \textbf{S\&C}, \textbf{S\&U}, and \textbf{P\&W}, defined by their interaction intent 
(see caption of Table~\ref{tab:android_results}), with the full mapping provided in Appendix~\ref{app_aw_class}.
For \textsc{WebArena}, we evaluate three lightweight yet representative domains (\textbf{Gitlab}, \textbf{Maps}, \textbf{Reddit}) to reduce computational cost.

\vspace{0.3em}
\noindent\textbf{Implementation Details.}
Following~\citep{sun2024genesis}, we use three VLM backbones
(InternVL2-4B/8B~\citep{chen2023internvl} and Qwen2-VL-7B~\citep{Qwen2VL}),
with \gpt~\citep{hurst2024gpt} for instruction synthesis and reward modeling.
All methods are trained under the same SFT setting with screenshots and
\atree{} inputs. We compare \ours{} against four representative data–synthesis
paradigms: \textsc{Zero-Shot} (CoT + M3A)~\citep{rawles2024androidworld}, \textsc{Task-Driven
Generation}~\citep{lai2024autowebglm}, \textsc{Self-Instruct}~\citep{wang2023selfinstruct},
and \textsc{OS-Genesis}~\citep{sun2024genesis}. \ours{} uses the same 1K-trajectory synthesis budget as \textsc{OS-Genesis}; the additional synthesis cost is bounded and tunable, with HD-MCTS controlled by the iteration budget and replay--refine capped by $F_{\max}$ with early stopping. Full training and efficiency details are provided in Appendix~\ref{app:exp_details}.

\footnotetext[2]{Baseline reproduction uses available resources including official checkpoints, released data, and our own reimplementations (see Appendix~\ref{app:exp_details}).}

\vspace{0.2em}
\noindent\textbf{Overall Performance.\footnotemark[2]}
\Cref{tab:android_results,tab:web_tasks} show that \ours{} achieves the
highest success rates across all backbones. Improvements are especially notable
in categories where existing methods perform poorly. For example, on
\textsc{AndroidWorld} \textbf{S\&C}, \textsc{OS-Genesis} reaches only \textbf{11.11\%} on InternVL2-4B and \textbf{0.00\%} on Qwen2-VL-7B,
while \ours{} attains \textbf{38.46\%} and \textbf{30.77\%}. Similar gains appear on \textbf{P\&W} (e.g.,
\textbf{1.85\%} vs.\ \textbf{9.88\%} on InternVL2-4B) and on \textsc{WebArena} domains such as \textbf{Gitlab}
(\textbf{7.94\%} vs.\ \textbf{22.73\%}). During baseline reproduction, we also observed that adding \textsc{Self-Instruct} data sometimes degraded \textsc{Task-Driven} performance due to instruction–execution misalignment and noisy supervision, further illustrating the sensitivity of GUI agents to data quality. Following \citep{sun2024genesis}, the GPT-4o–based M3A agent is reported only as a reference baseline. These representative cases highlight a consistent pattern: \ours{} 
substantially boosts performance precisely in interaction regimes that existing 
data-synthesis pipelines struggle with, leading to the highest overall success 
rates on both mobile and web environments. 
\footnotetext[3]{To save computation, we evaluate on a representative subset of \textsc{AndroidWorld};
    see Appendix~\ref{app_subset_details}.}

\vspace{0.2em}
\noindent\textbf{Component Analysis.\footnotemark[3]} We next analyze the effectiveness of the three components of \ours{}: (i) the alignment-based hardness metric, (ii) hardness-driven exploration, and (iii) alignment-guided refinement.

\begin{table}[t]
\centering
\scriptsize
\caption{\textbf{Metric comparison of action-level alignment.}}
\vspace{-8pt}
\renewcommand{\arraystretch}{1.2}
\setlength{\tabcolsep}{6pt}
\label{tab:metric_ablation}
\begin{tabularx}{\linewidth}{
>{\centering\arraybackslash}X
>{\centering\arraybackslash}p{0.25\linewidth}}
\toprule[1.5pt]
\textbf{Alignment Metric Definition in Eq.~\ref{eq:recall}} & \textbf{Success Rate (\%)} \\
\midrule[0.1pt]
\midrule[0.1pt]
\rowcolor[HTML]{E6F3FF}
$\tfrac{1}{|A_a|} \sum_{a \in A_a} \mathbf{1}\!\left[\exists\, b \in B_a:\, \mathrm{match}(a,b\,|\,s)\right]$
& \textbf{50.8} \\
\midrule[0.1pt]
$\tfrac{1}{|B_a|} \sum_{b \in B_a} \mathbf{1}\!\left[\exists\, a \in A_a:\, \mathrm{match}(a,b\,|\,s)\right]$
& 36.0 \\
\bottomrule[1.5pt]
\end{tabularx}
\vspace{-2em}
\end{table}

\vspace{0.2em}
\noindent\textbf{Analysis I. Hardness Metric Validation.}
We validate the hardness definition in Eq.~\ref{eq:hardness} through two complementary analyses: parameter sensitivity and metric substitution.

\begin{itemize}[leftmargin=1.5em]
\item \textbf{Parameter sensitivity.}
We vary $\epsilon$ and $\alpha$ in Eq.~\ref{eq:hardness} and observe consistent trends (Table~\ref{tab:metric_param_ablation}). 
Large $\epsilon$ flattens hardness differences, while small or large $\alpha$ distort the reward landscape and reduce
discriminative power. 
The default configuration ($\epsilon{=}0.01$, $\alpha{=}1$) offers a balanced
scaling that preserves sensitivity to semantic hardness while maintaining
stable reward behavior.

\begin{figure}[!b]
\centering
\begin{subfigure}[b]{0.98\linewidth}
    \centering
    \includegraphics[width=\linewidth]{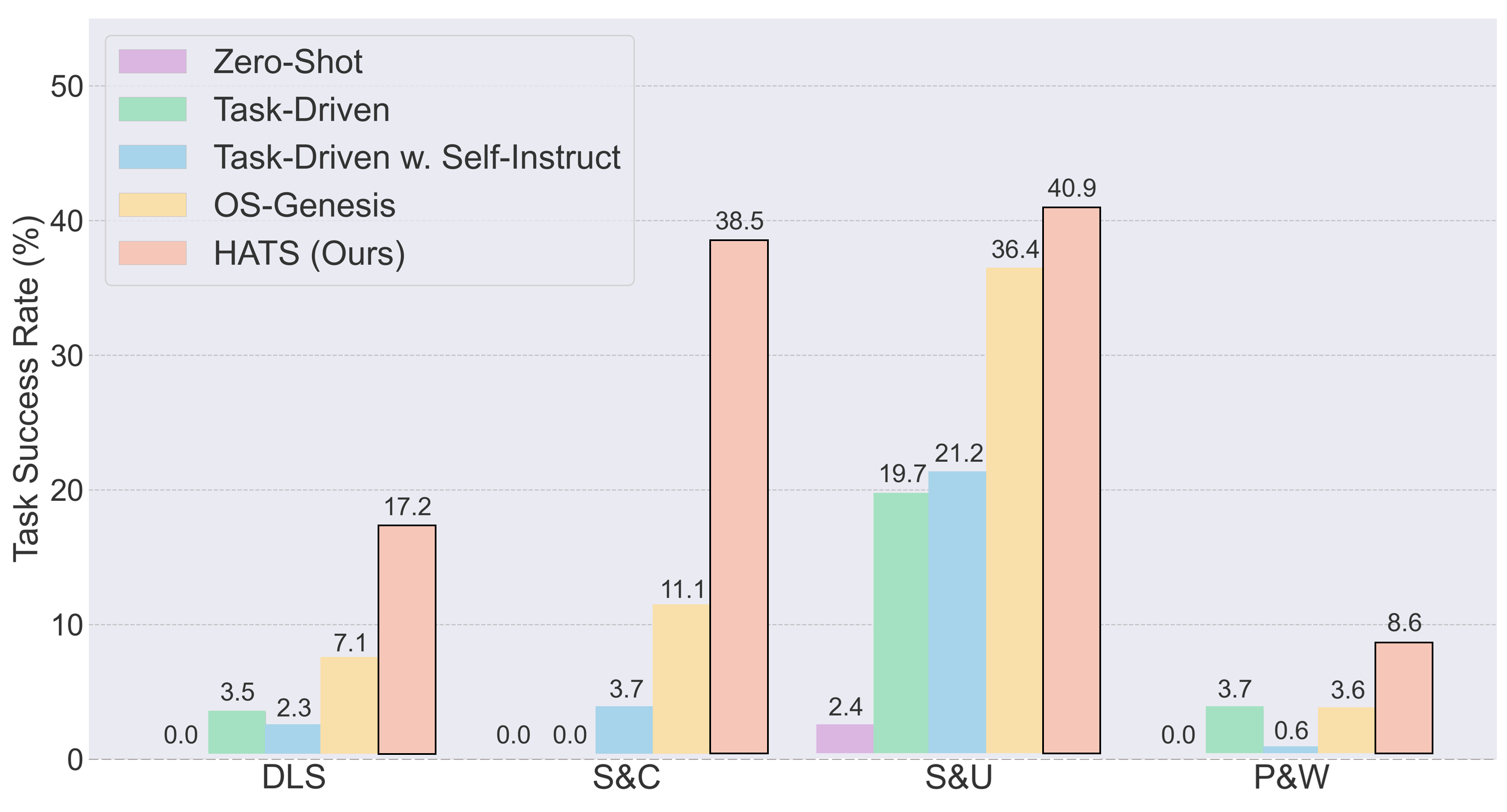}
    \caption{Performance comparison on \textsc{AndroidWorld} with InternVL2-4B.}
\end{subfigure}
\begin{subfigure}[b]{0.495\linewidth}
    \centering
    \includegraphics[width=\linewidth]{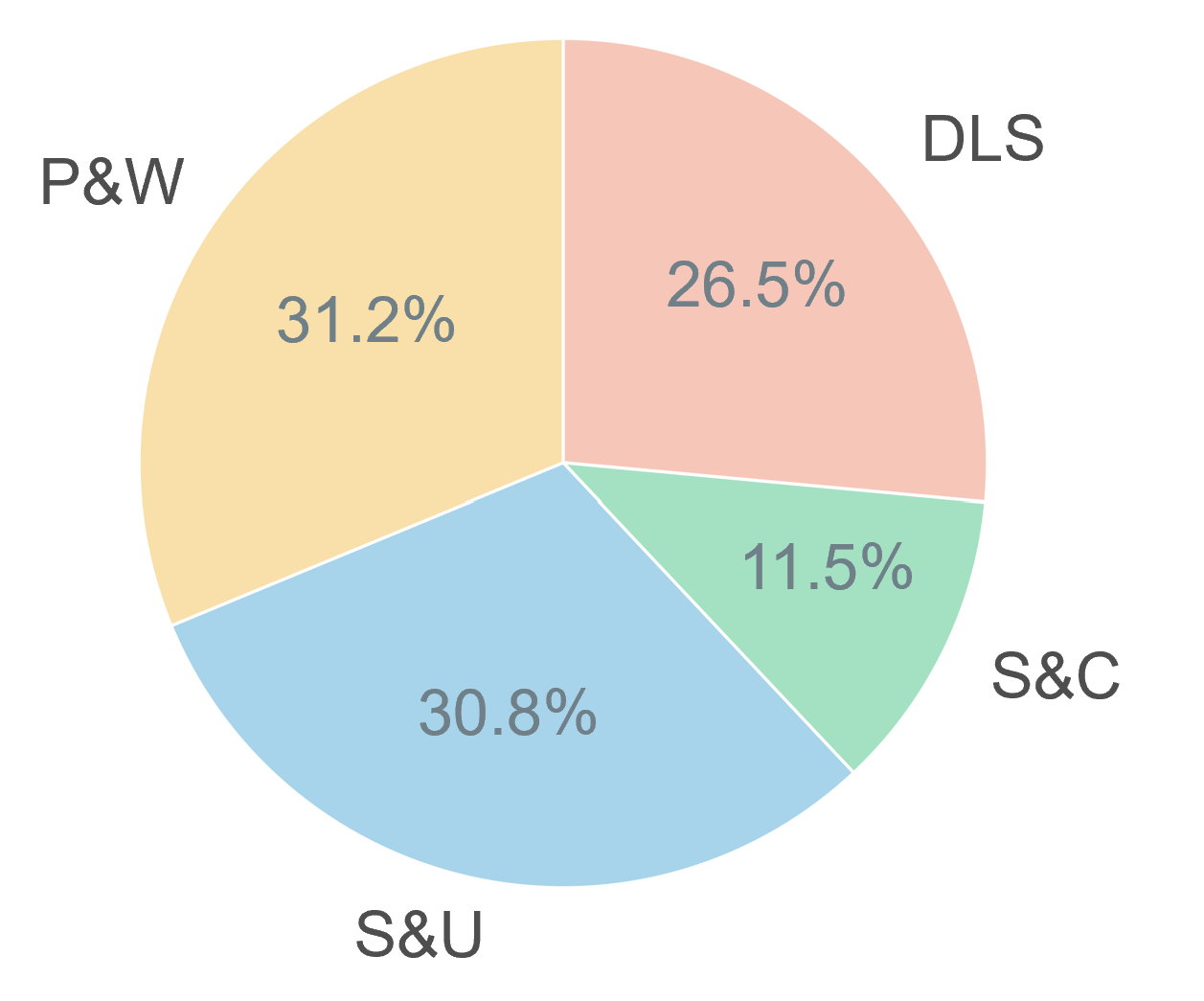}
    \caption{Distribution: Baseline \citep{wang2023selfinstruct}}
\end{subfigure}
\hfill
\begin{subfigure}[b]{0.495\linewidth}
    \centering
    \includegraphics[width=\linewidth]{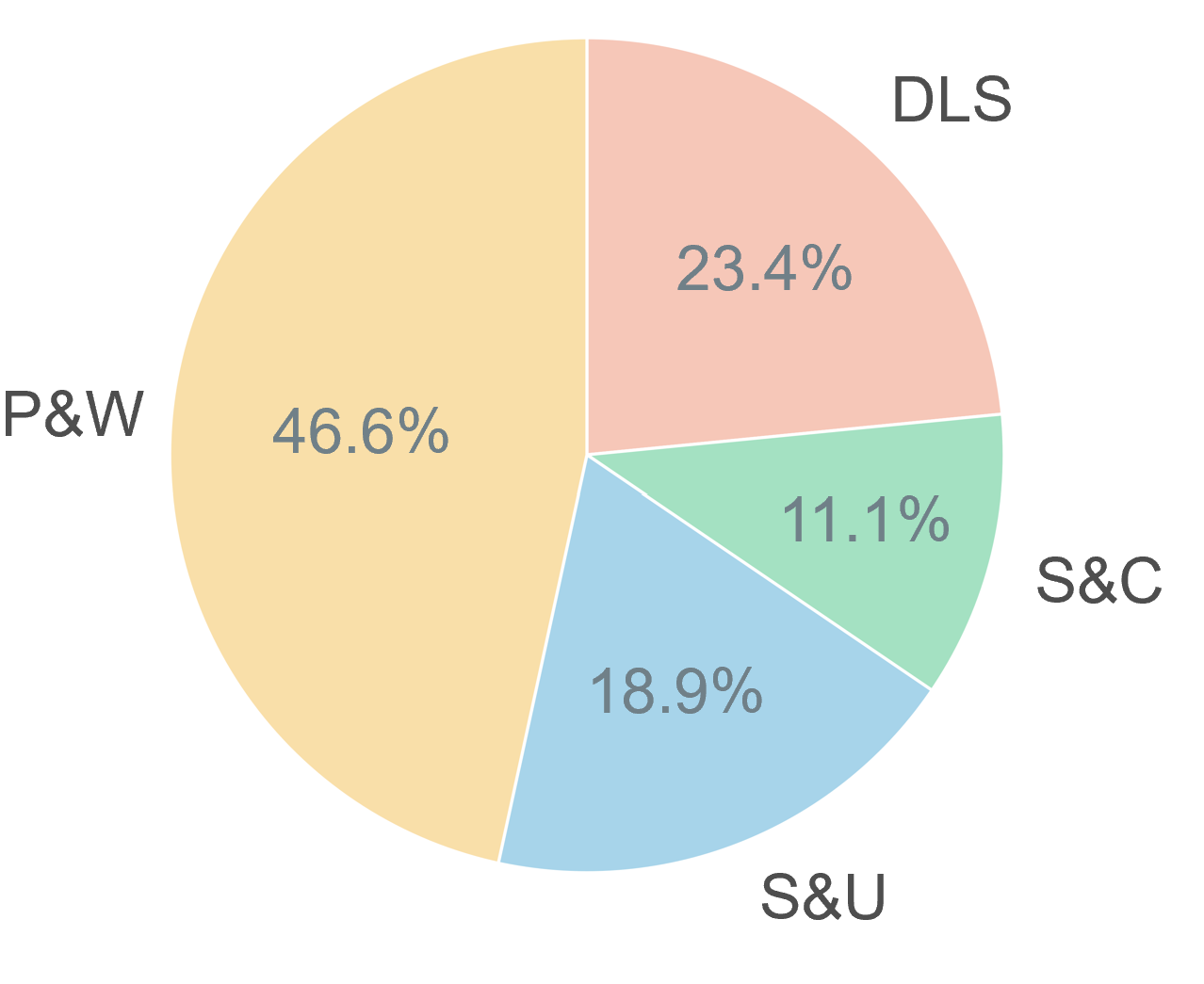}
    \caption{Distribution: Ours}
\end{subfigure}
\caption{\textbf{Performance comparison \& corpus composition.} Top: performance comparison on \textsc{AndroidWorld} using InternVL2-4B. 
Bottom: corpus composition by task category for the baseline (left) and \ours{} (right), where \ours{} increases exploration in categories that are empirically harder for existing methods (e.g., \textbf{P\&W}), while reducing exploration in easier 
categories (e.g., \textbf{S\&U}).
}
\label{fig:combined_three_plots}

\vspace{-2em}
\end{figure}

\item \textbf{Metric comparison.}
We substitute recall in Eq.~\ref{eq:recall} with an action-level precision
metric and derive an analogous hardness mapping. As shown in Table~\ref{tab:metric_ablation}, recall-based hardness correlates
substantially better with downstream performance: recall captures coverage of intended actions, which is the main failure mode in ambiguous GUI workflows; precision does not penalize missing steps.
\end{itemize}

\begin{figure}[t]
\centering

\begin{subfigure}[b]{0.495\columnwidth} 
\centering
\includegraphics[width=\textwidth]{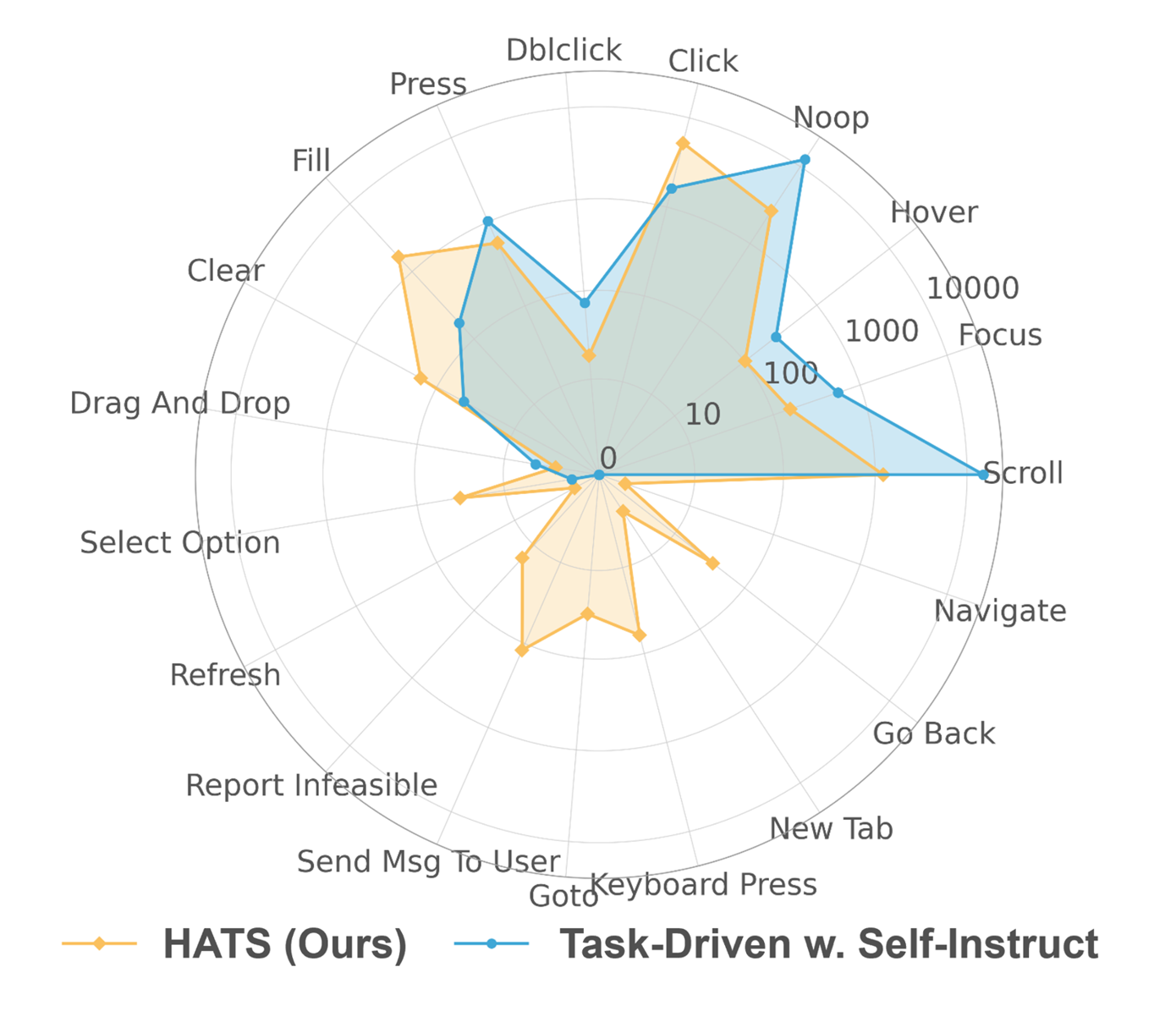}
\caption{\scriptsize{Anal. II: Action Type Distribution}}
\end{subfigure}
\hfill
\begin{subfigure}[b]{0.495\columnwidth}
\centering
\includegraphics[width=\textwidth]{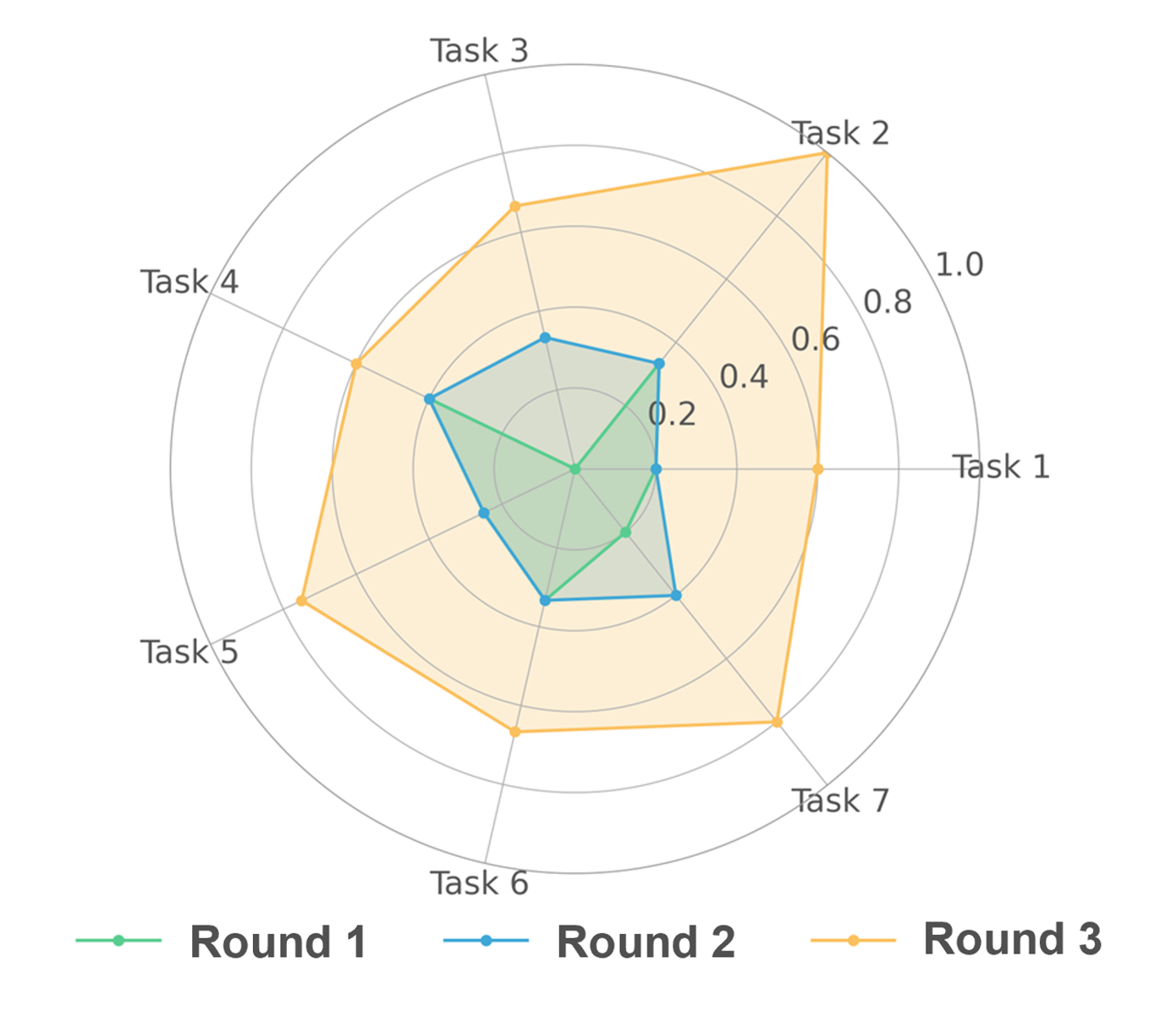}
\caption{\scriptsize{Anal. III: Refinement Effect}}
\end{subfigure}
\caption{
\textbf{Radar chart comparison.}
(a) Action-type distribution on \textsc{WebArena}, showing distributional differences
between the baseline and \ours{}.  
(b) Per-task view of multi-round refinement, where the shaded area expands
across rounds, indicating higher action-level recall and better synthesized
trajectories.
}
\label{fig:radar_combined}
\vspace{-1em}
\end{figure}

\vspace{2em}
\noindent\textbf{Analysis II. Effectiveness of Hardness-Driven Exploration.} We use two qualitative analyses to show that hardness-driven exploration
indeed shifts sampling coverage toward semantically valuable, ambiguity-rich regions.

\footnotetext[4]{OS-Genesis (10/2025) does not provide category-labeled training data; we thus use \textsc{Task-Driven w.\ Self-Instr.} for category-level comparison.}

\begin{figure}[h]
\centering
\includegraphics[width=0.95\columnwidth]{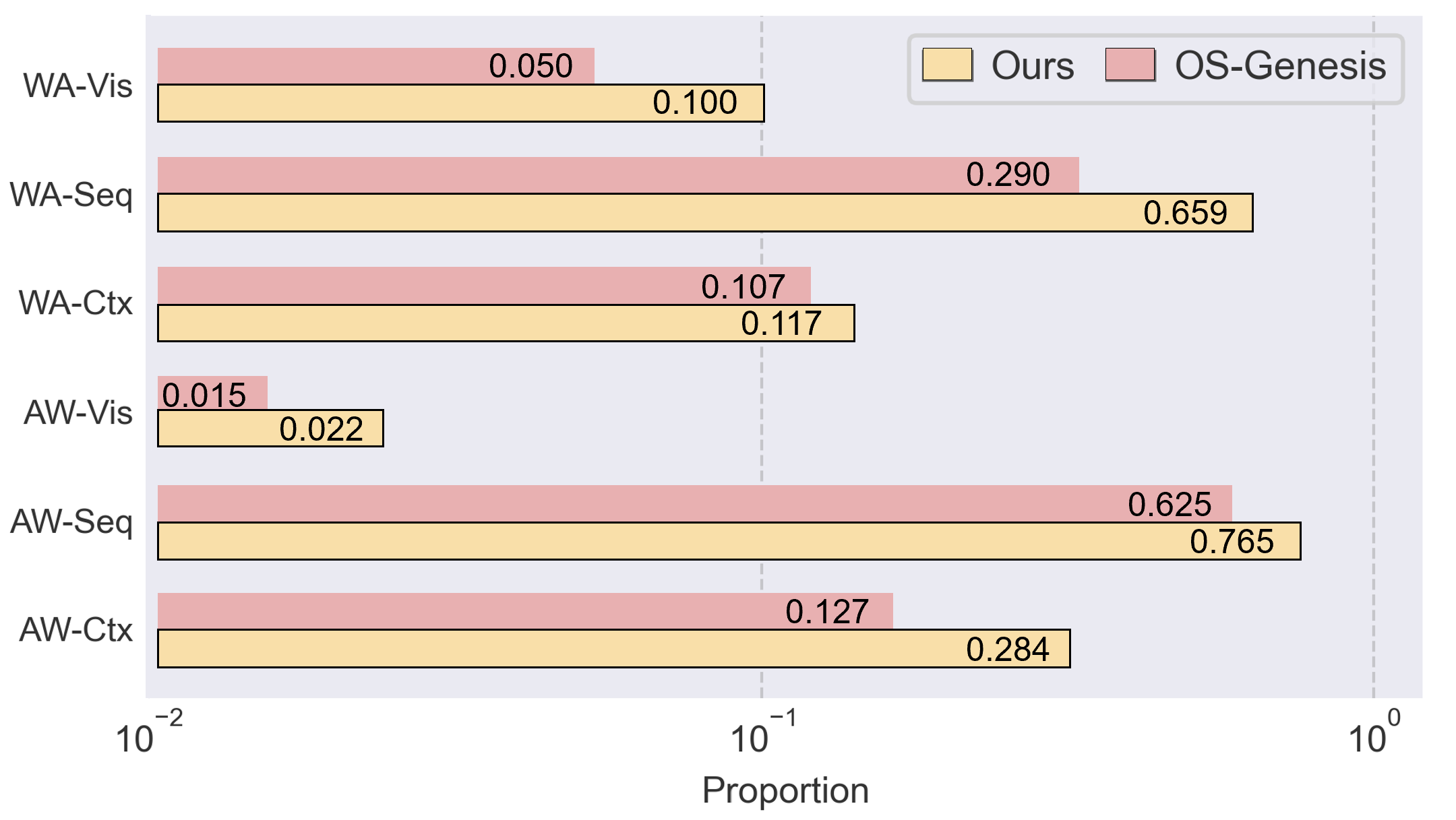} 
\caption{
\textbf{Semantic-ambiguity ratios at log scale.}
\ours{} vs.\ \textsc{OS-Genesis} across ambiguity types 
Vis, Seq, and Ctx on \textsc{AndroidWorld} (AW) and \textsc{WebArena} (WA). Bars correspond to AW/WA-Ctx/Seq/Vis pairs.
Trajectories may involve multiple ambiguity types, so categories are not mutually exclusive.
}
\label{fig:ambig_audit}
\vspace{-1em}
\end{figure}

\begin{itemize}[leftmargin=1.5em]
\item \textbf{Distributional coverage analyses.}
Fig.~\ref{fig:combined_three_plots}a reveals a clear pattern across
baselines: tasks in \textbf{S\&U} exhibit uniformly high success rates, whereas
\textbf{P\&W} tasks show consistently lower success, suggesting that \textbf{P\&W}
contains a larger fraction of semantically complex, multi-step workflows.
Motivated by this observation, we compare category proportions in the synthesized trajectories
(Fig.~\ref{fig:combined_three_plots} (b-c).  
Relative to the baseline\footnotemark[4], \ours{} increases the sampling weight of \textbf{P\&W} from
\textbf{31.2\%} to \textbf{46.6\%}, while reducing \textbf{S\&U} from \textbf{30.8\%} to \textbf{18.9\%}.  
This redistribution aligns with the empirical hardness indicated in
Fig.~\ref{fig:combined_three_plots}a: harder categories receive more
coverage, while easier ones are deemphasized. A similar shift is observed in the action-type distributions on \textsc{WebArena}
(Fig.~\ref{fig:radar_combined}a), which likely contributes to its improved performance.

\item \textbf{Semantics-focused auditing.} We further compare whether the final trajectories synthesized by \ours{} contain
more semantic ambiguity than those produced by \textsc{OS-Genesis}.  
Using a third-party LLM auditor (\gpt) following our rubric in Appendix \ref{app_subset_details}, each trajectory is independently evaluated for three
ambiguity types: context dependency, sequential dependency, and visual
ambiguity.  
As shown in Fig.~\ref{fig:ambig_audit}, a consistently larger fraction of
\ours{} trajectories receive ambiguity labels across all three categories,
indicating that hardness-driven exploration uncovers interactions where
language–action grounding is intrinsically challenging.
\end{itemize}

\vspace{0.4em}
\noindent\textbf{Analysis III. Effectiveness of Alignment-Guided Refinement.} 
We evaluate how multi-round replay--refinement improves instruction--execution alignment.
\label{sec:refine_effect}
\begin{itemize}[leftmargin=1.5em]
    \item \textbf{Overall effect.}  
    Table~\ref{tab:refine_ablation} shows that, refinement yields substantial gains: 
    the average action-level reconstruction recall $R(A,B)$ (Eq.~\ref{eq:recall}) 
    increases from \textbf{0.26} to \textbf{0.40}, 
    accompanied by a corresponding improvement in downstream success rate 
    (from \textbf{16.00\%} to \textbf{23.07\%}). 
    This indicates that iterative replay increasingly corrects mismatches and 
    strengthens overall instruction fidelity.

    \item \textbf{Single-case insight.}  
    A per-task view in Fig.~\ref{fig:radar_combined}b, using seven representative
    tasks (Appendix~\ref{app:refine_tasks}), exhibits the same steadily improving trend: 
    the radar area expands monotonically across rounds, showing that iterative correction 
    progressively fills missing context, resolves sequential inconsistencies, 
    and produces more executable task instructions for individual behaviors.
\end{itemize}

\begin{table}[h]
\centering
\scriptsize
\caption{\textbf{
Ablation on multi-round refinement on AndroidWorld.
}}
\vspace{-8pt}
\renewcommand{\arraystretch}{1}
\setlength{\tabcolsep}{5pt}
\label{tab:refine_ablation}
\begin{tabularx}{\linewidth}{
>{\centering\arraybackslash}p{0.40\linewidth}|
>{\centering\arraybackslash}p{0.20\linewidth}
>{\centering\arraybackslash}p{0.265\linewidth}}
\toprule[1.5pt]
\textbf{Refinement Rounds} & \textbf{$R_{\text{avg}}$} & \textbf{Success Rate (\%)} \\
\midrule[0.1pt]
\midrule[0.1pt]
One-Shot
& 0.26\,{\tiny -35.0\%} 
& 16.00\,{\tiny -30.6\%} \\

Round 1
& 0.19\,{\tiny -52.5\%} 
& 15.56\,{\tiny -32.6\%} \\

Round 2
& 0.29\,{\tiny -27.5\%} 
& 21.43\,{\tiny -7.1\%} \\

\rowcolor[HTML]{E6F3FF}
Round 3
& \textbf{0.40}\,{\tiny 0.0\%} 
& \textbf{23.07}\,{\tiny 0.0\%} \\

\bottomrule[1.5pt]
\end{tabularx}
\vspace{-1em}
\end{table}

\section{Conclusion}
\label{sec:conc}

We introduced \ours, a closed-loop framework that transforms instruction–execution feedback into a hardness signal to guide trajectory synthesis for GUI agents. 
Through the integration of a \textbf{hardness-driven exploration} module and an \textbf{alignment-guided refinement} module, the framework jointly enhances the diversity of semantic-ambiguous actions and the semantic consistency of generated data. 
By promoting exploration toward challenging yet informative interactions and ensuring alignment at the instruction–trajectory level, \ours produces higher-quality, more learnable datasets that enable GUI agents to generalize more robustly across real-world interfaces.


{
    \small
    \bibliographystyle{ieeenat_fullname}
    \bibliography{main}
}

\clearpage

\appendix

\section{Extended Related Works}
\label{app:related}
\noindent\textbf{Language-Driven GUI Automation.}
Large language models (LLMs) have become the engine behind a new wave of language-conditioned agents that perceive, reason, and act within graphical user interfaces~\citep{durante2024agent,feng2024far,wu2024oscopilot,zhou2026hiconagent}.
Early efforts retained frozen parameters and relied on prompt programming, cooperative agent collectives~\citep{wu2023autogen,sun2023corex,jia2024agentstore}, external tool execution~\citep{sun2024survey}, self-reflection loops~\citep{shinn2024reflexion,xu2024envisions,cheng2024vision}, and simulated world models~\citep{hu2023language,jin2024mmtom,yang2023appagent}.
A parallel line of work adapts model weights to acquire interface-specific skills: recognizing visual screen states~\citep{cheng2024seeclick,gou2024navigating,wu2024atlas} or hierarchical UI trees~\citep{OSWorld,zheng2024seeact}, emitting atomic actions such as click and type~\citep{chen2024guicourse}, and generalizing across web~\citep{yao2022webshop,deng2023mindweb}, desktop~\citep{kapoor2024omniact,niu2024screenagent}, and mobile domains~\citep{li2024androidcontrol,wang2024mobileagent}.
Related trends are also visible beyond GUI automation, where multimodal grounding, history-aware policy optimization, and sequential decision-making have been studied in media understanding and embodied agents~\citep{shao2023detecting,shao2024detecting,li2025cogvla,li2025semanticvla,li2026consisvla4d,li2026global,li2026optimusvla,zhu2026H-GAR,shao2025large}.
Collectively, these studies lay the groundwork for agents capable of non-trivial task execution in heterogeneous digital ecosystems.

\vspace{0.2em}
\noindent\textbf{Benchmark Suites and GUI Corpora.}
Robust agent training hinges on rich, varied interface data capturing both static layouts and dynamic behaviours~\citep{wu2024oscopilot,zeng2024agenttuning,pan2024autonomous}.
The \textsc{Rico} dataset introduced sequential screenshots and view hierarchies for mobile applications~\citep{deka2017rico}, whereas \textsc{MiniWob} presented fine-grained web interactions~\citep{shi2017world}.
Subsequent initiatives extended coverage: large-scale mobile GUI logs~\citep{rawles2023aitw,zhang2024aitz,lu2024guiodyssey,chai2024amex}, expansive web task collections~\citep{zheran2018miniwobplus,lu2024weblinx,murty2024nnetscape}, and curated desktop traces for instructional tasks~\citep{chen2024guicourse}.
Related data-centric efforts in embodied and structured interaction learning have likewise emphasized scalable action datasets and temporal consistency~\citep{lv2025spatial,huang2024scalable}.
These corpora supply the visual and structural context needed to ground language-conditioned policies.

\vspace{0.2em}
\noindent\textbf{Trajectory-Centric Supervision.}
End-to-end agent learning benefits from trajectory data composed of instructions, sub-goals, actions, and intermediate states~\citep{li2024androidcontrol,zhang2024agentohana,zheng2024agentstudio}.
Two major paradigms exist:
(i) \textbf{Task-driven} schemes recruit annotators or scripted bots to complete predefined tasks, producing high-fidelity but narrow-coverage demonstrations~\citep{li2024androidcontrol,lu2024weblinx};  
(ii) \textbf{Exploration-driven} pipelines allow autonomous agents to traverse environments and discover tasks through random walks~\citep{sun2024genesis,pahuja2025explorer}.
Recent embodied-agent studies similarly rely on trajectory abstraction, hierarchical refinement, memory augmentation, and skill composition to improve long-horizon action quality~\citep{li2025star,li2025cogvla,li2025semanticvla,zhu2026H-GAR,li2026consisvla4d,li2026global,li2026optimusvla}.
While cost-efficient, naive exploration gathers large numbers of semantic-intuitive actions while failing to capture rarer but more valuable \textbf{semantic-ambiguous actions}.
Consequently, existing corpora remain biased toward trivial behaviours and often contain incoherent instructions or execution errors, limiting generalization.
Alleviating this imbalance and improving semantic alignment directly motivates the present work.

\vspace{0.2em}
\noindent\textbf{Hardness-Aware Learning.}
The concept of hardness has been long studied in machine learning.
Curriculum and self-paced learning~\citep{bengio2009curriculum,kumar2010self} organize training from easy to hard samples, while automated curricula~\citep{graves2017automated} and hardness-aware exploration~\citep{florensa2018automatic,portelas2020automatic} dynamically adjust sample hardness.
In computer vision, hard-example mining techniques prioritize rare or ambiguous samples that improve model generalization~\citep{shrivastava2016training}, and related robustness research further highlights the need to focus on adversarially sensitive, hard-to-evaluate, or domain-shifted cases~\citep{gao2021maximum,gao2022fast,shao2019multi,shao2017deep}.
Yet existing GUI trajectory synthesis pipelines seldom incorporate hardness-aware mechanisms~\citep{sun2024genesis,pahuja2025explorer}.
Our method addresses this gap by transforming instruction--execution misalignment into a hardness signal to identify and prioritize under-represented, high-value semantic-ambiguous actions.

\section{Experimental Details}
\label{app:exp_details}

\subsection{Evaluation Benchmarks}
\label{app_aw_class}
We evaluate our method on two representative GUI benchmarks: \textsc{AndroidWorld}~\citep{rawles2024androidworld} (mobile) and \textsc{WebArena}~\citep{zhou2024webarena} (web).

\subsubsection{Environment Setup}
\label{app:bench_config}

\begin{itemize}[leftmargin=1.5em]
\item \textsc{AndroidWorld}~\citep{rawles2024androidworld}.
This mobile-focused benchmark is executed in Android emulators and contains diverse user-oriented tasks such as app navigation, form filling, and in-app purchases, capturing realistic GUI layouts and interaction dependencies.
We conduct all mobile experiments using Android Virtual Devices (AVD) configured with Pixel 6 hardware profiles and the Tiramisu system image (API Level 33), following the official setup guidelines.

\item \textsc{WebArena}~\citep{zhou2024webarena}.
This large-scale web interaction benchmark consists of functional websites, including email clients, calendars, and e-commerce platforms, presenting challenges due to dynamic page layouts and varied interaction semantics.
For web-based evaluation, we employ the AgentLab framework~\citep{drouin2024workarena,chezelles2025browsergym} to standardize experiment execution and management, evaluating on three representative domains: \textbf{Gitlab} (180 tasks), \textbf{Maps} (109 tasks), and \textbf{Reddit} (106 tasks), which together cover diverse patterns of web interaction including repository management, location-based services, and social media platforms.
\end{itemize}

\begin{figure*}[!t]
\centering
\begin{minipage}{0.98\textwidth}
\begin{algorithm}[H]
\caption{\textbf{HD-MCTS}: Hardness-Driven Exploration with Alignment-Guided Refinement.}
\label{alg:hdmcts_final}
\begin{algorithmic}[1]
\Require Environment $\mathcal{E}$; depth limit $T_{\max}$; UCB constant $C$; thresholds $R_{\min}, F_{\max}$; hardness parameters $\epsilon, \alpha$
\For{\textbf{each iteration} \Comment{Each iteration generates one training sample}}
  \State Retrieve root node $v_0 \!\gets\! \textsc{GetOrCreateNode}(a_0, s_0)$; \quad $\mathcal{P}\!\gets\!\emptyset$
  \State $v \!\gets\! v_0$; \quad $s \!\gets\! s_0$

  \Comment{\textbf{Selection}: traverse by UCB until a frontier or depth limit (measured by $|\mathcal{P}|$)}
  \While{$v$ is fully expanded \textbf{and} $|\mathcal{P}| < T_{\max}$}
    \State $a^\star \!\gets\! \arg\max\limits_{a\in\mathcal{A}(v)} \Big[ Q(v,a) + C\sqrt{\tfrac{\ln(N(v){+}1)}{N(v,a){+}1}} \Big]$
    \State $s' \!\gets\! \mathcal{E}(s,a^\star)$
    \State $\mathcal{P} \!\gets\! \mathcal{P} \cup \{(a^\star, s')\}$; \quad $v \!\gets\! \textsc{GetChildNode}(a^\star,s')$; \quad $s \!\gets\! s'$
  \EndWhile

  \Comment{\textbf{Expansion}: add a previously unexpanded action if $|\mathcal{P}|{<}T_{\max}$}
  \If{$v$ \textbf{is not} fully expanded \textbf{and} $|\mathcal{P}| < T_{\max}$}
    \State choose previously unexpanded action $a_e\in\mathcal{A}(v)$; \quad $s_e\!\gets\!\mathcal{E}(s,a_e)$
    \State $u\!\gets\!\textsc{CreateNode}(a_e,s_e)$; \quad $\mathcal{P}\!\gets\!\mathcal{P}\cup\{(a_e,s_e)\}$; \quad $v\!\gets\!u$; \quad $s\!\gets\!s_e$
  \EndIf

  \Comment{\textbf{Simulation}: \underline{rollout} $\rightarrow$ \underline{selection} $\rightarrow$ \underline{instruction synthesis} $\rightarrow$ \underline{replay} $\rightarrow$ \underline{refine} $\rightarrow$ \underline{reward computation}}
  \State $\bar{\mathcal{P}} \!\gets\! \textsc{Rollout}(\mathcal{E}, v, s, \mathcal{P}, T_{\max})$
  \Comment{Extend path $\mathcal{P}$ from current node until reaching $T_{\max}$}
  \State $A \!\gets\! \textsc{SelectSubsequence}(\bar{\mathcal{P}})$ \Comment{Select a semantically coherent sub-trajectory from the path}
  \State $I \!\gets\! \textsc{SynthesizeInstruction}(A)$ \Comment{Instruction synthesis from $A$}
  \State $\mathbb{A} \!\gets\! \textbf{false}$; \quad $F \!\gets\! 0$ \Comment{$\mathbb{A}$: alignment flag}
  \Repeat
    \State $B \!\gets\! \textsc{ExecuteInstruction}(I)$ \Comment{Replay $I$ from the original start state of $A$}
    \State $R \!\gets\! \textsc{AlignmentScore}(A,B)$ \Comment{Action-level alignment, defined by Eq.~\ref{eq:recall}}
    \If{$R \ge R_{\min}$ \textbf{ and } \textsc{Executable}$(B)$}
      \State $\mathbb{A} \!\gets\! \textbf{true}$
    \Else
      \State $I \!\gets\! \textsc{RefineInstruction}(I, A, B)$ \Comment{Diagnose misalignment and refine}
      \State $F \!\gets\! F + 1$
    \EndIf
  \Until{$\mathbb{A}$ \textbf{ or } $F \ge F_{\max}$}

  \Comment{\textbf{Hardness reward}: convert alignment into exploration signal}
  \State $r \!\gets\! (R + \epsilon)^{-\alpha}$ \Comment{Eq.~\ref{eq:hardness}; lower alignment $\Rightarrow$ higher hardness}

  \Comment{\textbf{Emission}: keep only aligned pairs for training}
  \If{$\mathbb{A}$}
    \State \textsc{StoreSample}$(I, B, r)$ \Comment{Emit verified instruction–trajectory with hardness score}
  \EndIf

  \Comment{\textbf{Backpropagation}: update edge values along the selected path $\mathcal{P}$}
  \For{each $(a_i,s_i)\in\mathcal{P}$}
    \State $N(a_i)\!\gets\!N(a_i){+}1$
    \State $Q(a_i)\!\gets\!Q(a_i){+}\tfrac{r - Q(a_i)}{N(a_i)}$
  \EndFor
\EndFor
\end{algorithmic}
\end{algorithm}
\end{minipage}
\vspace{-1em}
\end{figure*}

\subsubsection{Task Distribution and Categorization}
\noindent\textbf{AndroidWorld Task Categories.}
We organized AndroidWorld's 116 evaluation tasks into four functional categories based on the primary application domain. Table~\ref{tab:app_categorization} provides the complete mapping from applications to categories. The four categories are defined as:
\begin{itemize}[leftmargin=1.5em]
    \item \textbf{DLS (Daily Life \& Services):} entertainment, media, and lifestyle apps
    \item \textbf{S\&C (Social \& Communication):} messaging and social interaction apps
    \item \textbf{S\&U (System \& Utility):} core system functions and utilities
    \item \textbf{P\&W (Productivity \& Work):} task management and productivity tools.
\end{itemize}
This categorization allows us to analyze performance across different levels of interaction complexity and semantic domains. The training corpus comprises \textbf{1,000 trajectories} collected through our synthesis pipeline, while the evaluation uses the full set of \textbf{116 benchmark tasks} from \textsc{AndroidWorld}.

\begin{table}[h]
\centering
\footnotesize
\caption{\textbf{Application categorization for AndroidWorld.}}
\vspace{-6pt}
\renewcommand{\arraystretch}{1.1}
\setlength{\tabcolsep}{4pt}
\begin{tabularx}{\linewidth}{%
  >{\centering\arraybackslash}X
  >{\centering\arraybackslash}X|
  >{\centering\arraybackslash}X
  >{\centering\arraybackslash}X}
\toprule
\rowcolor[HTML]{D9D9D9}
\textbf{Application} & \textbf{Category} &
\textbf{Application} & \textbf{Category} \\
\midrule
Calculator    & P\&W & VLC            & DLS \\
Calendar      & P\&W & Markor         & P\&W \\
Clock         & S\&U & RetroMusic     & DLS \\
Contacts      & S\&U & Gallery        & DLS \\
Dialer        & S\&U & Tasks          & P\&W \\
Settings      & S\&U & SMSMessenger  & S\&C \\
Chrome        & P\&W & Expense        & DLS \\
Files         & P\&W & Joplin         & P\&W \\
Google Tasks  & P\&W & Camera         & S\&U \\
Keep Notes    & P\&W & Broccoli       & DLS \\
Draw          & P\&W & Audio Recorder & P\&W \\
Launcher      & S\&U & OsmAnd         & S\&C \\
Markup        & P\&W & OpenTracks     & P\&W \\
\bottomrule
\end{tabularx}
\label{tab:app_categorization}
\vspace{-1em}
\end{table}

\subsubsection{Evaluation Protocol}
\noindent\textbf{Success Criteria.}
We adopt the official evaluation protocol from both benchmarks. Success is determined by \textbf{exact state matching}: an agent completes a task successfully if and only if the final environment state exactly matches the ground-truth target state specified by the benchmark. Partial completion or near-miss outcomes are counted as failures.

\noindent\textbf{Execution Constraints.}
Each task is subject to the maximum step limit defined by the respective benchmark. For AndroidWorld, we enforce per-task step budgets as specified in the original evaluation suite. For WebArena, we similarly respect the official constraints for each domain.

\vspace{0.2em}
\noindent\textbf{Retry Policy.} We retry executions only in cases of \textbf{environment-level failures} (e.g., AVD unresponsiveness, API timeouts, network errors). Task failures due to agent limitations (incorrect actions, semantic misunderstandings) are \textbf{not} retried and are recorded as failures in the final evaluation.

\subsection{Model Configuration and Training}
For instruction synthesis and reward modeling, we employ \gpt~\citep{hurst2024gpt}.  
Two VLM backbones are used for agent training:  
\begin{itemize}[leftmargin=1.5em]
    \item \textbf{InternVL2-4B/8B}~\citep{chen2023internvl}: general-purpose models without GUI-specific pretraining.
    \item \textbf{Qwen2-VL-7B-Instruct}~\citep{Qwen2VL}: an instruction-tuned model with agentic reasoning capabilities.
\end{itemize}

\subsubsection{Training Hyperparameters}
\label{app:training_details}

Table~\ref{tab:training_hyperparams} summarizes the training configuration for each VLM backbone. All models are trained for \textbf{15 epochs} with a global batch size of \textbf{128} (achieved via gradient accumulation) on \textbf{8$\times$NVIDIA H100 80GB GPUs}.

\begin{table}[h]
\centering
\footnotesize
\caption{\textbf{Training hyperparameters for VLM backbones.}}
\vspace{-6pt}
\renewcommand{\arraystretch}{1.1}
\setlength{\tabcolsep}{4pt}

\begin{tabularx}{\linewidth}{%
  >{\centering\arraybackslash}X
  >{\centering\arraybackslash}X
  >{\centering\arraybackslash}X
  >{\centering\arraybackslash}X}
\toprule
\rowcolor[HTML]{D9D9D9}
\textbf{Model} & \textbf{Learning Rate} & \textbf{Batch Size} & \textbf{Epochs} \\
\midrule
InternVL2-4B & 4e-5 & 128 & 15 \\
InternVL2-8B & 4e-5 & 128 & 15 \\
Qwen2-VL-7B & 2e-6 & 128 & 15 \\
\bottomrule
\end{tabularx}
\label{tab:training_hyperparams}
\vspace{-1em}
\end{table}

\vspace{0.2em}
\noindent\textbf{Optimization and Regularization.}
We use the \textbf{AdamW} optimizer~\citep{loshchilov2019decoupledweightdecayregularization} with a weight decay of \textbf{0.05} for all models. For Qwen2-VL-7B, we additionally apply gradient clipping with a maximum gradient norm of \textbf{1.0} to stabilize training. All models employ a \textbf{cosine annealing} learning rate schedule with a linear warmup phase covering \textbf{3\%} of total training steps (warmup ratio = 0.03).

\vspace{0.2em}
\noindent\textbf{Distributed Training Infrastructure.}
Training employs the \textbf{DeepSpeed}~\citep{rasley2020deepspeed} framework for efficient distributed optimization. The effective batch size of 128 is achieved with a per-device batch size of \textbf{2} and \textbf{8} gradient accumulation steps. We use \textbf{bfloat16} (bf16) mixed precision to enable full fine-tuning of all model parameters while maintaining stable gradient updates and memory efficiency.

\vspace{0.2em}
\noindent\textbf{Data Preprocessing.}
To accelerate training and inference, all input screenshots are resized to a resolution of \textbf{460$\times$1024} pixels while preserving aspect ratio. This resolution balances visual detail retention with computational efficiency.

\vspace{0.2em}
\noindent\textbf{Inference Configuration.}
During evaluation, we use \textbf{greedy decoding} (temperature = 0) to ensure deterministic and reproducible results. For AndroidWorld, we impose no explicit length constraints on model generation. For WebArena, due to longer observation sequences from complex web pages, we set \texttt{max\_model\_len=16384} to prevent out-of-memory errors while accommodating typical task contexts. For all experiments, we report results using the \textbf{final checkpoint} saved at the end of training (epoch 15).

\subsection{Baselines}
We compare \ours{} against four representative data-synthesis paradigms, all using consistent multimodal inputs (\atree and screenshots):

\begin{itemize}[leftmargin=1.5em]
    \item \textbf{Zero-Shot (CoT + M3A)}~\citep{rawles2024androidworld}: a GPT-based agent guided by Chain-of-Thought prompting~\citep{wei2022chain}, without additional training.
    \item \textbf{Task-Driven Generation}~\citep{lai2024autowebglm}: the model receives predefined tasks and screenshots to synthesize trajectories.
    \item \textbf{Self-Instruct}~\citep{wang2023selfinstruct}: automatically generates new task instructions to expand trajectory diversity.
    \item \textbf{OS-Genesis}~\citep{sun2024genesis}: a reverse task-synthesis pipeline that explores environments first and generates instructions post hoc, with instruction refinement and reward filtering.
\end{itemize}

\vspace{0.2em}
\noindent\textbf{Baseline Reproduction Details.}
To ensure rigorous comparison, we adopted specific reproduction protocols for different benchmarks.
On \textsc{AndroidWorld}, we evaluated \textsc{OS-Genesis} using their officially released weights and evaluation code on our identical AVD setup, while the \textsc{Task-Driven} and \textsc{Self-Instruct} pipelines were reproduced by ourselves following the methodologies described in \citep{sun2024genesis}.
On \textsc{WebArena}, consistent with \textsc{AndroidWorld}, we reproduced the \textsc{Task-Driven} and \textsc{Self-Instruct} baselines based on the descriptions in \citep{sun2024genesis}. However, the official evaluation code for \textsc{OS-Genesis} is not publicly available. Our attempts to reproduce its results yielded significantly lower performance than reported in the original paper. This discrepancy likely stems from environmental differences—such as updates to the underlying LLM API (e.g., gpt-4o) and subtle variations in browser state simulation—that are difficult to align without a standardized evaluation framework.
To avoid concerns about unfairly underestimating \textsc{OS-Genesis}, we report its original paper results in all main comparisons. Notably, our method still outperforms these baseline scores by a large margin, demonstrating that the observed gains are not affected by minor variances in reproduction.

\subsection{Trajectory-Based Training}
All methods use \textbf{Supervised Fine-Tuning} (SFT) over synthesized trajectories with two complementary objectives:
\[
\mathcal{L}_{1} = - \sum_{t_i \in \mathcal{T}} \log \left( p_{\theta}(\ell \mid s, h_i, c) \cdot p_\theta(a \mid s, h_i, c, \ell) \right)
\]
\vspace{-1em}
\[
\mathcal{L}_{2} = - \sum_{t_i \in \mathcal{T}} \log p_\theta(a \mid s, c, \ell)
\]
where $s$ is the multimodal state, $h_i$ the task instruction, $c$ the context, $\ell$ the subgoal, and $a$ the predicted action.  
We use 1K trajectories for all methods except Self-Instruct (1.5K), with an average of 6.4 interaction steps per trajectory. ReAct-style reasoning~\citep{yao2023react} is adopted for interpretability.

\subsection{Experimental Setup for Component Analysis}
\label{app_subset_details}
To balance computational efficiency with representativeness, component analyses were conducted on a curated subset of \textsc{AndroidWorld} comprising 20 tasks. These tasks span diverse functional domains—such as social communication, productivity, and utilities—and vary in interaction depth (3–12 steps), ensuring broad coverage of real-world agent behaviors. For each analysis, we employed identical rollout trajectories and model backbones (InternVL2-4B) to isolate the effect of each proposed component. For instance, in Analysis I (Hardness Metric Validation), we computed hardness values across multiple $(\epsilon, \alpha)$ configurations using the same trajectories, comparing both recall- and precision-based formulations under consistent conditions.

\vspace{-0.3em}
\section{Prompt Summary}
\label{app:prompt_summary}
We summarizes all prompt templates used in the paper:
\begin{itemize}[leftmargin=1.5em]

\item \textbf{Instruction Generation Prompt.}
For converting trajectories into natural-language instructions, we use a prompt that enforces step grounding, UI-action consistency, 
and minimal hallucination. 
This template ensures that generated instructions remain aligned with the underlying hard actions and faithfully reflect the agent’s behavior.

\item \textbf{Alignment Verification Prompt.}
To evaluate whether a synthesized instruction is semantically compatible with a rollout, we employ an alignment-checking prompt that 
compares action semantics, reasoning flow, and intended outcomes. 
This template is used in the alignment-aware refinement stage to filter out misaligned trajectories. 

\item \textbf{Semantic-Ambiguity Detection Prompt.}
To identify context-, sequential-, and visually driven hard actions, we employ a structured ambiguity-classification prompt. 
This prompt guides the model to analyze trajectories and determine whether they contain any of the three types of semantic ambiguity defined in the main paper. 

\end{itemize}
Together, these prompts define the interface between the agent, the trajectory, and the verifier. For completeness and reproducibility, the full prompt templates are provided in the following pages.

\vspace{-0.3em}
\section{Instruction Refinement Examples}
\label{app:refine_tasks}

Table~\ref{tab:refinement_tasks} shows the task instructions for the seven representative tasks in Fig.~\ref{fig:radar_combined}b, refined over three rounds by the \textbf{Alignment-Guided Refinement} module. As the process progresses, instructions are systematically enriched with critical semantic details—such as precise UI element properties (e.g., “clickable and focusable”), required navigation steps (e.g., “scroll down if the target is off-screen”), and explicit interaction modalities—leading to a consistent and monotonic increase in action-level recall. This demonstrates how structured, grounded refinement directly enables more accurate and executable agent behavior.

\vspace{-0.3em}
\section{Illustrative Examples}

\begin{figure*}[!t]
\centering
\includegraphics[width=\linewidth]
  {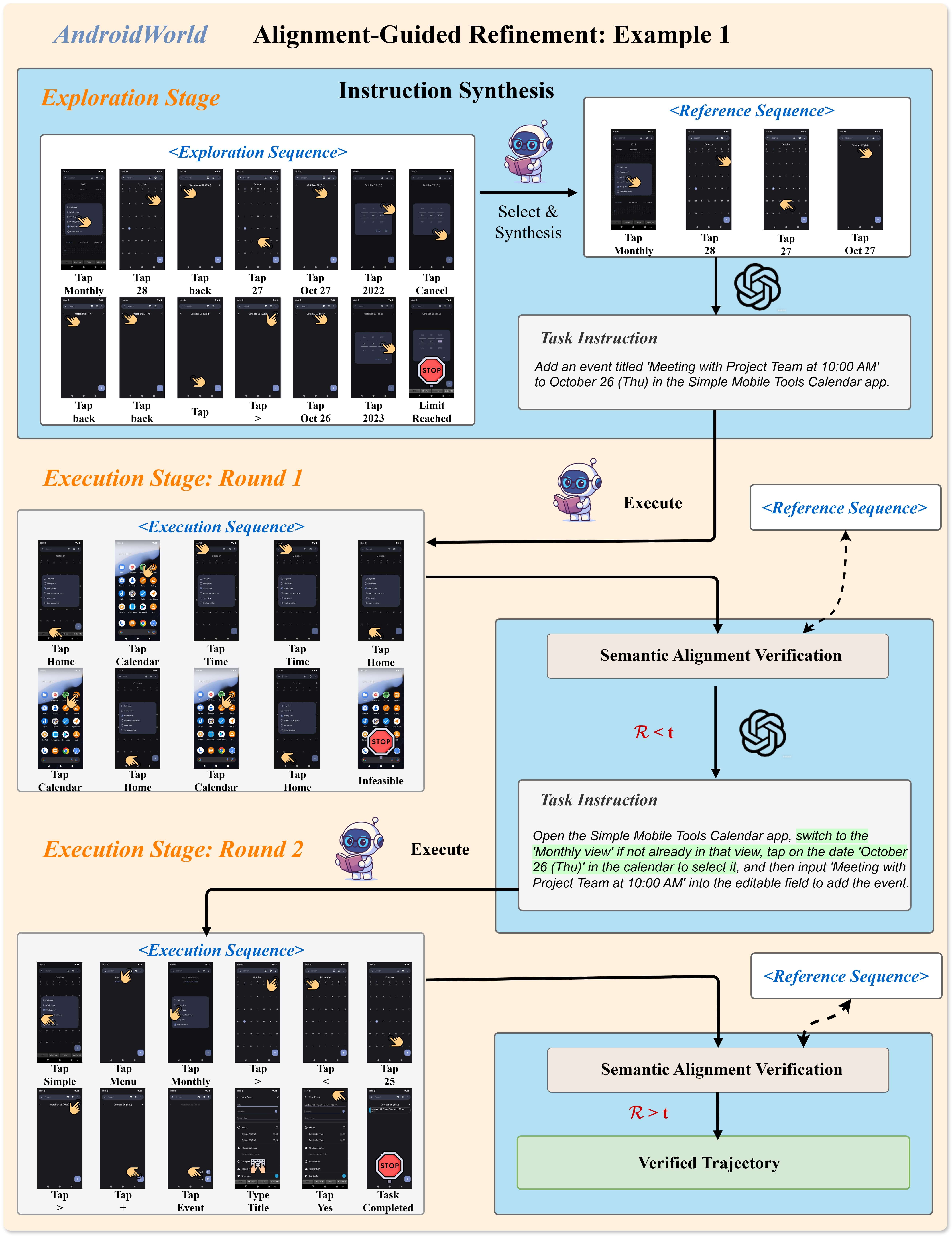}
\phantomcaption 
\label{fig:example_1}
\end{figure*}

\begin{figure*}[!t]

  \centering
  \includegraphics[width=\linewidth]
  {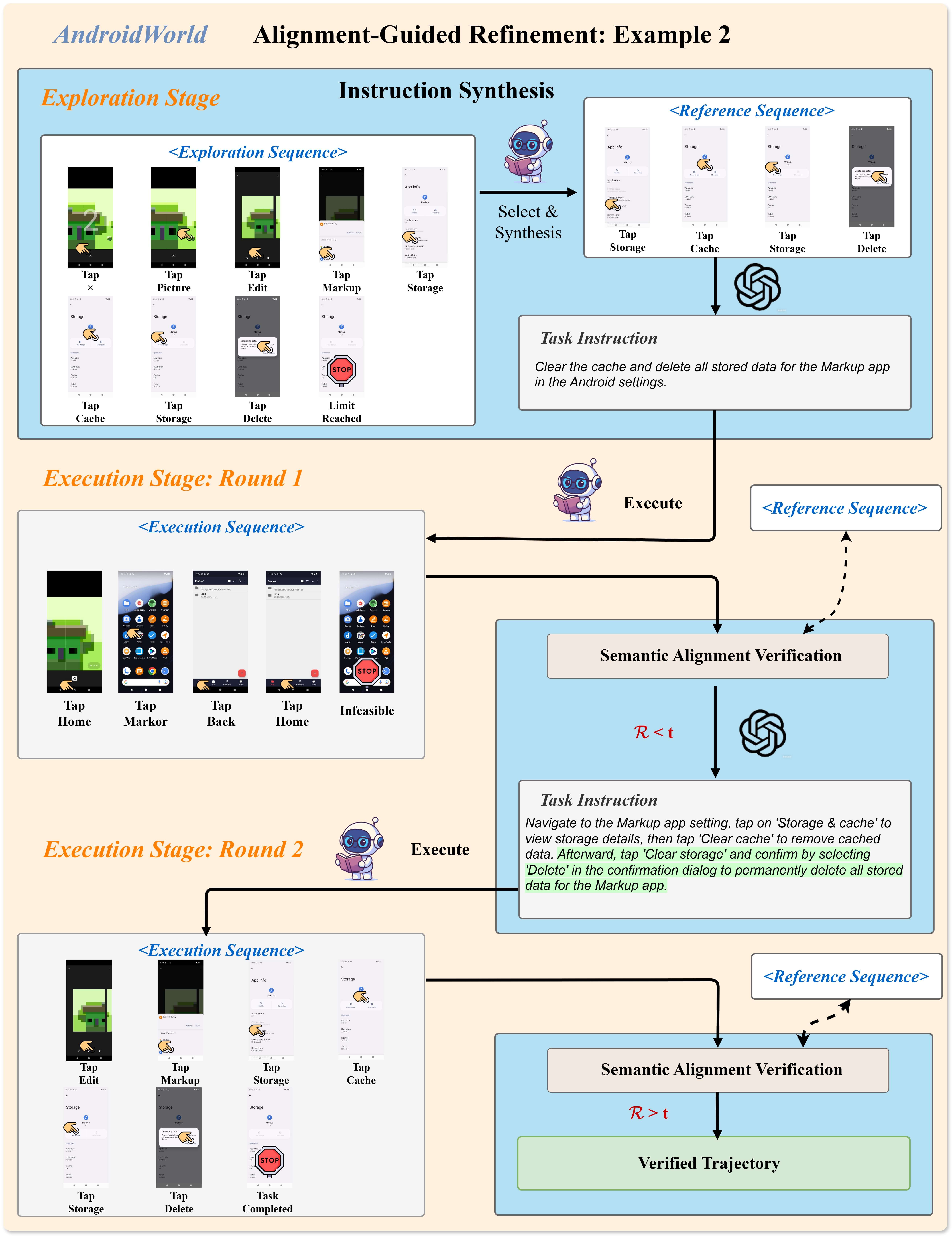}
\phantomcaption 
  \label{fig:example_2}
\end{figure*}

\begin{figure*}[!t]

  \centering
  \includegraphics[width=\linewidth]
  {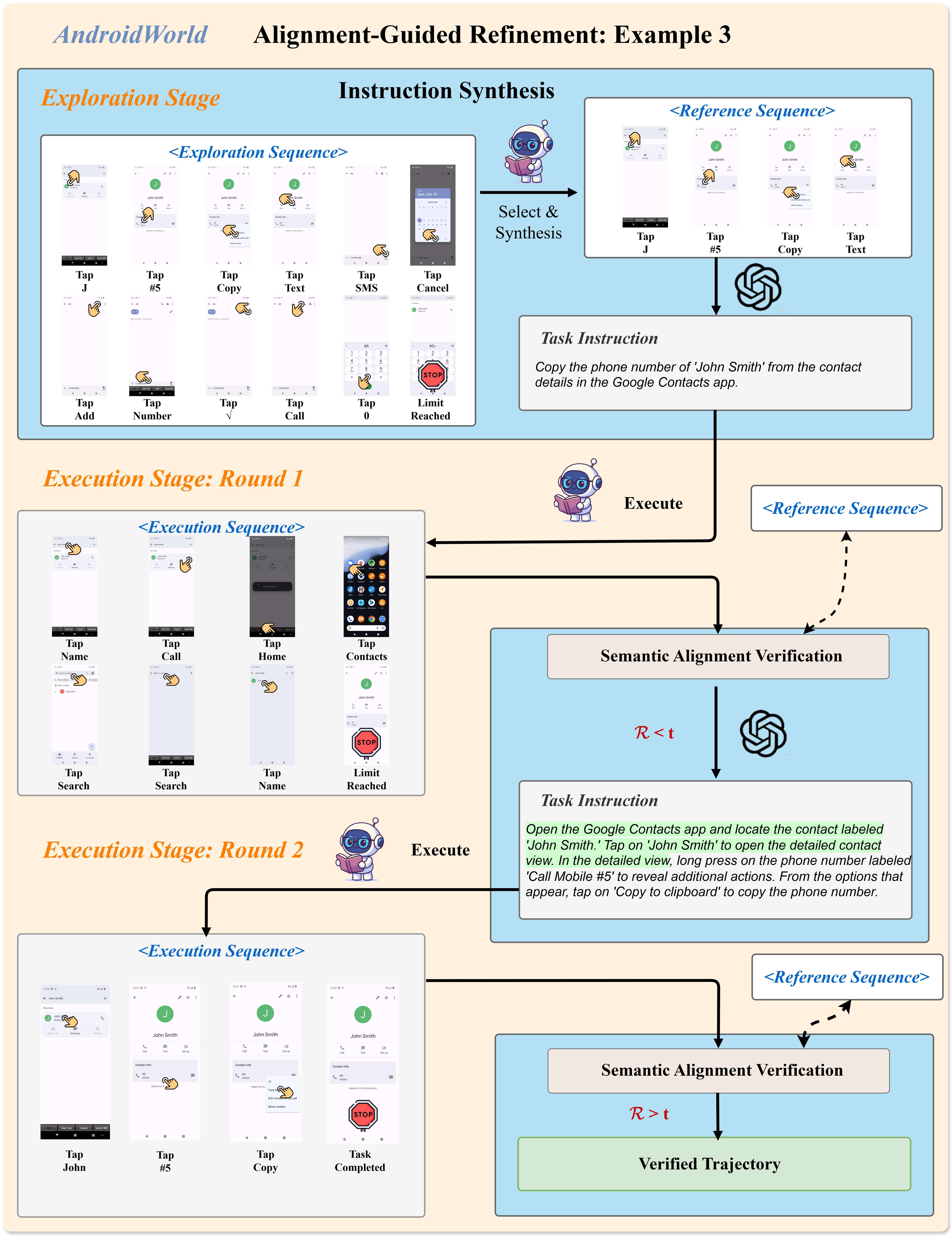}
  \phantomcaption 
  \label{fig:example_3}
\end{figure*}

\begin{figure*}[!t]

  \centering
  \includegraphics[width=\linewidth]
  {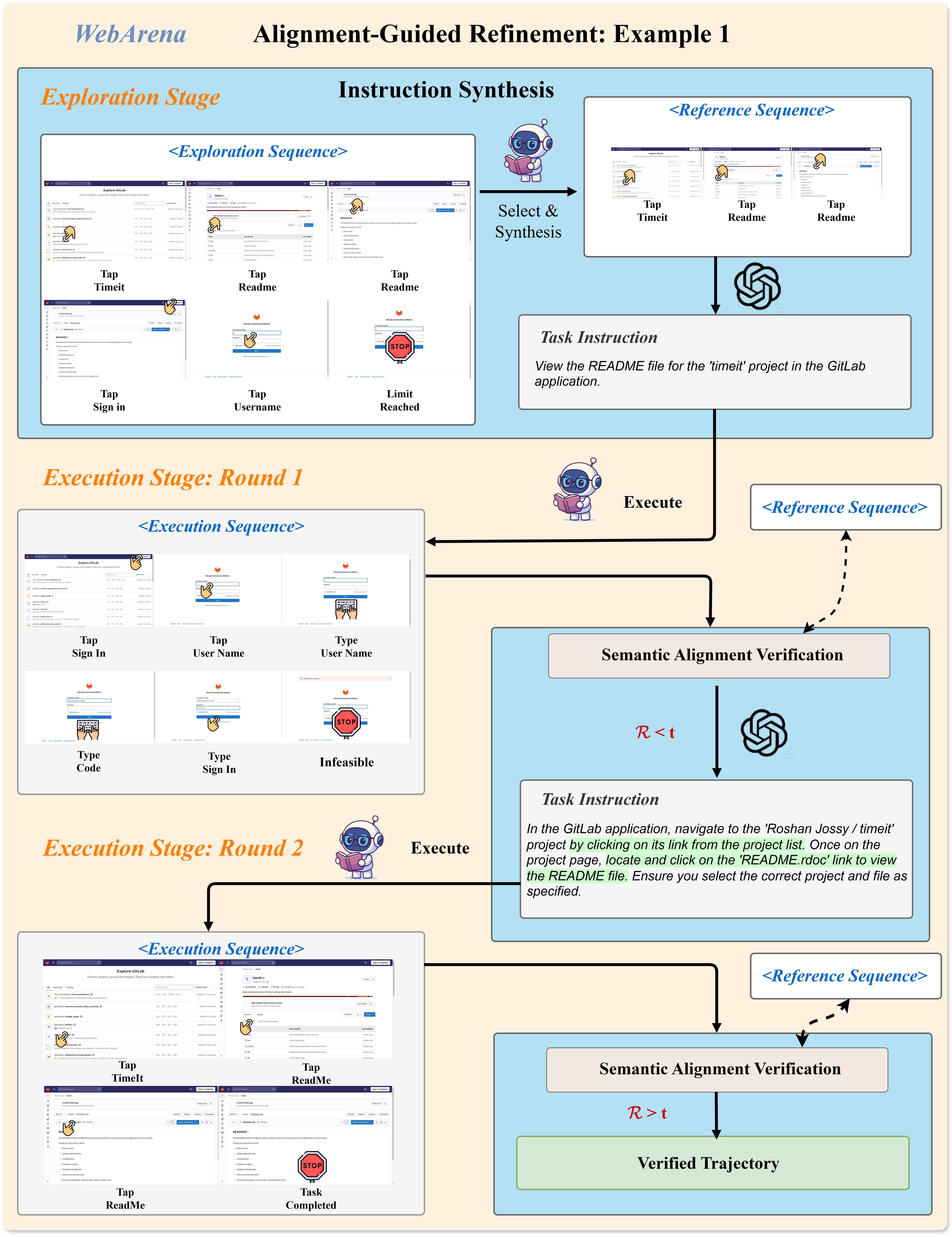}
  \phantomcaption 
  \label{fig:example_4}
\end{figure*}

In Figures~\ref{fig:example_1} to~\ref{fig:example_4}, we present four illustrative cases of alignment-guided refinement, drawn from \textsc{AndroidWorld} (three) and \textsc{WebArena} (one). These examples are deliberately selected for their clarity and representativeness, enabling clear visualization of each stage in our pipeline: exploration, reference selection, instruction synthesis, execution, and refinement. The initial \textbf{Exploration Sequence} is typically noisy and repetitive, obscuring the underlying task intent. For example, in Example~1, the agent repeatedly navigates across multiple pages with redundant actions, making it impossible to infer the target date. From this raw trajectory, we extract a compact \textbf{Reference Sequence} that retains only goal-relevant steps, effectively distilling intent—such as ``perform operations for October 26.'' This refined sequence guides the generation of an initial task instruction, which the agent executes to produce an \textbf{Execution Sequence}. While this first attempt shows improved logical structure, it often lacks critical details, resulting in failure. For instance, the agent may omit the specific date due to an underspecified instruction. Through our \textbf{Refine} phase, we enrich the instruction semantically by aligning it with the reference sequence, enabling the agent to recover missing context. In Example~1, the refined instruction explicitly states, ``Navigate to October 26, then click 'Confirm,''' leading to successful task completion. These cases demonstrate that our alignment-guided refinement mechanism transforms ambiguous, exploratory traces into precise, executable instructions, enabling agents to reliably complete complex, real-world GUI tasks.

\clearpage

\clearpage

\clearpage

\clearpage

\begin{onecolumn}
\begin{small}
\begin{longtable}{p{0.065\textwidth} c p{0.8\textwidth}}
\caption{Evolution of synthesized instructions across three refinement rounds for the tasks in Figure~\ref{fig:radar_combined}b.} \label{tab:refinement_tasks} \\
\toprule
\textbf{Task ID} & \textbf{Round} & \textbf{Synthesized Instruction} \\
\midrule
\endfirsthead

\multicolumn{3}{c}%
{{\bfseries \tablename\ \thetable{} -- continued from previous page}} \\
\toprule
\textbf{Task ID} & \textbf{Round} & \textbf{Synthesized Instruction} \\
\midrule
\endhead

\midrule
\multicolumn{3}{r}{{Continued on next page}} \\
\bottomrule
\endfoot

\bottomrule
\endlastfoot
\multirow{3}{*}{\textbf{Task 1}} 
& 1 & In the Audio Recorder app, rename the audio recording titled '05-06-2025 06.14.28PM' to 'Meeting\_Notes\_05-06-2025'. First, double-click the text field containing '05-06-2025 06.14.28PM' to select it. Then, double-click the 'Copy' option from the edit menu. Double-click the date and time field again to ensure focused editing, and double-click the 'Paste' option from the edit menu. Finally, input 'Meeting\_Notes\_05-06-2025' into the text field to set the new name. \\
& 2 & In the Audio Recorder app, rename the audio recording titled '05-06-2025 06.14.28PM' to 'Meeting\_Notes\_05-06-2025'. First, locate the text field containing '05-06-2025 06.14.28PM' in the renaming prompt and double-click it to select the text. Then, double-click the 'Copy' option from the edit menu above the selected text. Next, double-click the same text field again to ensure it is focused for editing. After that, click the 'Paste' option from the edit menu to paste the copied text. Finally, input 'Meeting\_Notes\_05-06-2025' into the text field to set the new name, and click the 'Save' button to confirm the changes. \\
\rowcolor[HTML]{E6F3FF} & 3 & In the Audio Recorder app (package name: com.dimowner.audiorecorder), properly set up the renaming interface to change the audio recording titled '05-06-2025 06.14.28PM' to 'Meeting\_Notes\_05-06-2025'. Start by double-clicking the text field containing '05-06-2025 06.14.28PM' to select it in the renaming prompt. Then, double-click the 'Copy' option from the edit menu above the selected text to copy it. Double-click the text field to refocus it if necessary. Click the 'Paste' option from the edit menu to paste the copied text. Finally, manually input 'Meeting\_Notes\_05-06-2025' into the text field for renaming and press the 'Save' button to confirm the change. Ensure that you are on the renaming dialog screen directly accessing the desired recording and follow the sequence precisely. \\
\cmidrule{1-3}

\multirow{3}{*}{\textbf{Task 2}} 
& 1 & Delete the audio records 'Record-10' and 'Record-4' in the Audio Recorder app. \\
& 2 & In the Audio Recorder app, first tap on the bookmark icon next to 'Record-10' to select it, then double-tap on 'Record-4' to select it. Finally, tap on the trash bin icon in the top-right corner to delete the selected records. \\
\rowcolor[HTML]{E6F3FF} & 3 & In the Audio Recorder app, first tap on the bookmark icon next to 'Record-10' to select it. Then scroll down if necessary to ensure 'Record-4' is visible. Once visible, double-tap on 'Record-4' to select it by quickly tapping twice on the bookmark icon next to 'Record-4'. After both 'Record-10' and 'Record-4' are selected, tap on the trash bin icon in the top-right corner to delete these selected records. \\
\cmidrule{1-3}

\multirow{3}{*}{\textbf{Task 3}} 
& 1 & Sort the audio records by date (oldest) in the Audio Recorder app. \\
& 2 & In the Audio Recorder app (com.dimowner.audiorecorder), first tap on the filter icon to access sorting options. Then select 'By date (oldest)' to sort the audio records. After sorting, long-click on 'Record-11' to select it. \\
\rowcolor[HTML]{E6F3FF} & 3 & In the Audio Recorder app (com.dimowner.audiorecorder), tap on the filter icon, which is clickable and focusable, to access sorting options. Ensure you select 'By date (oldest)' from the sorting options to sort the audio records correctly. After sorting, locate 'Record-11' on the screen and long-click on it to select it. \\
\cmidrule{1-3}

\multirow{3}{*}{\textbf{Task 4}} 
& 1 & Disable all calendar notifications in the Simple Mobile Tools Calendar Pro app by navigating through the settings menu. \\
& 2 & Disable all calendar notifications in the Simple Mobile Tools Calendar Pro app by navigating to the 'Settings' menu, selecting 'Customize notifications' under the 'Reminders' section, toggling off the 'All Calendar notifications' option, and verifying that notifications are fully disabled. \\
\rowcolor[HTML]{E6F3FF} & 3 & Disable all calendar notifications in the Simple Mobile Tools Calendar Pro app by following these steps: 1) Tap on the 'Settings' icon located in the top-right corner of the app. 2) Select 'Customize notifications' under the 'Reminders' section. 3) Locate the toggle switch labeled 'All Calendar notifications' and ensure it is turned off. 4) Tap on the 'Calendar' icon to access additional notification settings. 5) Select 'Notifications' to return to the app info screen and verify that all calendar notifications are fully disabled. \\
\cmidrule{1-3}

\multirow{3}{*}{\textbf{Task 5}} 
& 1 & Explore Google's privacy and safety resources by navigating through the Privacy Policy and Safety Centre pages in the Chrome app. \\
& 2 & In the Chrome app, explore Google's privacy and safety resources by first tapping on the 'Privacy Policy' option in the Google Privacy \& Terms menu. Then, double-tap on the 'Privacy Principles' link within the Technologies page. Next, double-tap on the 'Safety Centre' link to navigate to the Safety Centre page. Finally, tap on the 'Safety Centre' logo to refresh or reload the page. \\
\rowcolor[HTML]{E6F3FF} & 3 & In the Chrome app, begin by navigating to the home screen if you are not already there, and open the Chrome app by tapping its icon. Once inside Chrome, access the Google Privacy \& Terms menu by tapping on 'Settings' and then 'Privacy and security'. Scroll down until you find the 'Privacy Policy' option and tap on it. Next, navigate to the Technologies page and double-tap on the 'Privacy Principles' link. Then, proceed to the Safety Centre page by double-tapping on the 'Safety Centre' link. Finally, refresh the Safety Centre page by tapping on the 'Safety Centre' logo. \\
\cmidrule{1-3}

\multirow{3}{*}{\textbf{Task 6}} 
& 1 & Navigate through the sidebar in the Joplin app to explore different sections like 'All notes', 'Notebooks', and 'Trash'. \\
& 2 & Open the sidebar menu in the Joplin app by tapping the 'Sidebar' icon. Once the sidebar is visible, scroll left if necessary to make sure all options are accessible. Tap the 'Back' arrow in the top-left corner to return to the previous screen, confirming each stage of navigation. \\
\rowcolor[HTML]{E6F3FF} & 3 & In the Joplin app, start by ensuring the main screen is visible. First, tap the 'Sidebar' icon, ensuring it is a clickable UI element with the content description 'Sidebar'. If the sidebar does not appear fully open or options aren't visible, scroll left on the scrollable UI element within the sidebar. Once you have verified the sidebar is open and accessible, identify the 'Back' arrow; it should be a visible, clickable UI element in the top-left corner of the screen. Tap the 'Back' arrow to return to the previous screen. Confirm each action by checking that the intended screen or menu appears after each tap or scroll. \\
\cmidrule{1-3}

\multirow{3}{*}{\textbf{Task 7}} 
& 1 & In the Tasks app, switch to the 'Default list', mark two 'New Task' entries as completed, and ensure a task is created using the 'Create new task' button by double tapping. \\
& 2 & In the Tasks app (org.tasks), first tap on the 'Default list' dropdown to select the 'Default list' if it's not already active. Then, locate the '+' icon with the 'Create new task' description and double tap it to create a new task. Next, scroll through the task list and find two tasks labeled as 'New Task'. Tap on the checkbox next to each 'New Task' entry to mark them as completed. Ensure the list is scrolled up or down as necessary to reveal the tasks and 'Create new task' button. \\
\rowcolor[HTML]{E6F3FF} & 3 & In the Tasks app (org.tasks), start by tapping on the 'Default list' dropdown to ensure the 'Default list' is active. If it's not already active, tap it once to select it. Next, locate each 'New Task' entry in the task list and tap the checkbox next to them to mark them as completed. Tap on the checkbox, and ensure you have scrolled, if necessary, to bring the tasks into view. Once this is done, locate the '+' icon described as 'Create new task', tap once to create a new task, then tap on the 'Save' icon with the 'Save' description to save the task. Make sure the sequence is followed as described to achieve the user goal. \\

\end{longtable}
\end{small}
\end{onecolumn}

\clearpage

\begin{promptbox}[Prompt for Input Field Content Generation]
\footnotesize
Generate appropriate input content for the input field marked with a green bounding box 
and labeled as number {index} in the screenshot. Your generated content should be suitable 
for the usage scenario shown in this screenshot and represent what real users would typically 
enter in daily use. Your response should contain only the text that needs to be entered into 
the input field.
 
\end{promptbox}

\phantomsection
\label{prompt:instr_gen}
\begin{promptbox}[Prompt for Trajectory-to-Instruction Conversion (Android)]
\footnotesize
You are an expert specializing in inferring specific user tasks based on changes observed 
in mobile phone screenshots within an interaction trajectory. I will provide you with an 
interaction trajectory containing:\\

1. Action per Step: One of click, double\_click, long\_click, scroll\_up, scroll\_down, 
   scroll\_left, scroll\_right, input. If the action is input, the input text is provided. 
   Each non-scroll action includes the target element's attributes 
   (e.g., content\_description, text, resource\_id).\\

2. Screenshots: A pre-action screenshot for each step (with a green bbox and step number) 
   and one final post-trajectory screenshot. Pay close attention to the element inside the 
   green bbox and all UI changes between consecutive screenshots.\\

3. Package Name: The package\_name of the active app (use it to infer the app's common name).\\

Note: In Sub-Instruction, Analysis, Knowledge, and Task-Instruction, do not use the 
numeric marker from the green bbox; refer to elements by their visible labels or roles 
(e.g., "the Settings icon", "the field labeled Username").\\

Part 1: Analysis and Step Selection\\
- Analyze the entire trajectory (all actions and screenshot changes).\\
- Select a logically coherent subsequence that completes a clear user task.\\
- Exclude redundant/irrelevant/irrational steps.\\

Part 2: Output Generation
Produce five parts (lists must match the original step count where specified):\\
- Sub-Instruction (List[str]): For each original step, write a concise, actionable instruction 
  grounded in the observed UI, including concrete identifiers (filenames, times, visible text).\\
- Analysis (List[str]): For each original step, reason about likely next actions based on 
  the new UI state.\\
- Knowledge (List[str]): For each original step, describe the general functionality of the 
  interacted element, inferred from before/after screenshots (1–2 sentences).\\
- Selected-Step-ID (List[int]): The indices of the steps you chose for the coherent subsequence, 
  in chronological order.\\
- Task-Instruction (str): A single actionable instruction that captures the user goal 
  of the selected subsequence. Explicitly mention the inferred app name and any crucial specifics.\\

You must return only a JSON dictionary in the following format:\\
\{\\
  "Sub-Instruction": List[str],\\
  "Analysis": List[str],\\
  "Knowledge": List[str],\\
  "Selected-Step-ID": List[int],\\
  "Task-Instruction": str\\
\}\\

The trajectory information will be provided below:
'{trajectory\_information}'\\

RETURN ME THE DICTIONARY I ASKED FOR.

\end{promptbox}
\clearpage

\begin{promptbox}[Android Action Reasoning \& Execution Prompt]
\footnotesize
\textbf{Role Definition}
You are an Android operation AI that fulfills user requests through precise screen interactions. 
Current and annotated screenshots are provided.\\

\textbf{Action Catalog (STRICT JSON FORMAT)}\\
1. Status Operations:\\
   - Task Complete: { "action\_type": "status", "goal\_status": "complete" }\\
   - Task Infeasible: { "action\_type": "status", "goal\_status": "infeasible" }\\
2. Information Actions:\\
   - Answer Question: { "action\_type": "answer", "text": "<answer\_text>" }\\
3. Screen Interactions:\\
   - Tap Element: { "action\_type": "click", "index": <visible\_index> }\\
   - Long Press: { "action\_type": "long\_press", "index": <visible\_index> }\\
   - Scroll: { "action\_type": "scroll", "direction": "up|down|left|right", 
               "index": <optional\_container\_index> }\\
4. Input Operations:\\
   - Text Entry: { "action\_type": "input\_text", "text": "<content>", 
                   "index": <text\_field\_index> }\\
   - Keyboard Enter: { "action\_type": "keyboard\_enter" }\\
5. Navigation:\\
   - Home Screen: { "action\_type": "navigate\_home" }\\
   - Back Navigation: { "action\_type": "navigate\_back" }\\
6. System Actions:\\
   - Wait Refresh: { "action\_type": "wait" }\\

\textbf{Current Objective}\\
User Goal: {task\_goal}\\

\textbf{Execution Context}\\
Action History:\\
{history}\\

Visible UI Elements (only interact with visible=true):\\
{ui\_elements}\\

\textbf{Core Strategy}\\
1) Path Optimization: prioritize app drawer for app launch; always use input\_text for 
   text fields; verify visibility before interacting; scroll containers (if possible) 
   before full-screen scrolls.\\
2) Error Handling: switch approach after > 1 failures; use scroll to reveal off-screen 
   targets; try opposite scroll if needed.\\
3) Information Tasks: use answer for questions; verify data freshness.\\
4) Expert Techniques: {knowledge}\\

\textbf{Response Format (STRICT)}\\
Reasoning: [step-by-step analysis covering visibility checks, history effectiveness, 
            alternatives, scroll considerations]\\
Action: [single JSON action from the catalog]\\

Generate response:

\end{promptbox}

\begin{promptbox}[Step Summary Generation Prompt]
\footnotesize
Goal: \{task\_goal\}\\

Before screenshot elements:
\{before\_ui\_elements\}\\

After screenshot elements:
\{after\_ui\_elements\}\\

Action: \{action\}
Reasoning: \{reasoning\}\\

Provide a concise single-line summary (< 50 words) comparing screenshots and action outcome. \\
Include: intended effect; success/failure; key info for subsequent steps; critique if 
action/reasoning was flawed; any important data to remember across apps. \\
For answer or wait (no screen change), assume success.\\

Summary:

\end{promptbox}

\clearpage
\phantomsection
\label{prompt:align_verify}
\begin{promptbox}[Matching Exploration vs.\ Agent Trajectories Prompt(Alignment Verification)]
\footnotesize
Objective: Determine which steps from the reference Exploration Trajectory were matched by 
the GUI Agent's trajectory. Report the count, matched exploration step indices, and the 
corresponding agent step indices.\\

Input:\\
1) Task Instruction:\\
{task\_instruction}\\
2) Exploration Trajectory (numbered "Step i"):\\
{exploration\_trajectory\_information}\\
3) GUI Agent Trajectory (numbered "Step i"):\\
{gui\_agent\_trajectory\_information}\\

Task: Compare trajectories; identify each exploration step matched by one or more agent steps; \\
list matched exploration indices; list all agent step indices contributing to any match; \\
count unique matched exploration steps.\\

Matching Rule: A match occurs if the agent performs the equivalent action or achieves the same \\
sub-goal. Multiple agent steps mapping to the same exploration step count as one toward the \\
match total. Still include all contributing agent step indices.\\

Output (JSON only):\\
\{\\
  "match\_num": <int>,\\
  "matched\_exploration\_id": [<int>, ...],\\
  "matched\_gui\_agent\_id": [<int>, ...]\\
\}\\

Constraints: match\_num == len(matched\_exploration\_id); sort both index lists ascending.\\

\end{promptbox}

\begin{promptbox}[Refining Task Instruction Prompt]
\footnotesize
Role: You refine a given task instruction so a GUI Agent's execution more closely 
matches a reference exploration trajectory.\\

Inputs:\\
1) Initial Task Instruction:\\
{task\_instruction}\\
2) Exploration Trajectory (ground-truth sequence):\\
{exploration\_trajectory\_information}\\
3) GUI Agent Trajectory (taken steps):\\
{gui\_agent\_trajectory\_information}\\
4) Matched Exploration Steps (low-level instructions the agent achieved):\\
{matched\_low\_level\_instructions}\\
5) Matched GUI Agent Steps (reasoning/action steps corresponding to matches):\\
{matched\_gui\_agent\_steps}\\

Your Task: Identify exploration steps not matched by the agent; analyze agent deviations; 
rewrite the task instruction to guide execution toward the full exploration sequence.\\

Refinement Guidelines: Add specifics from unmatched steps (element texts/descriptions, 
action types like long\_click or scroll\_up, required input text); clarify sequence; address 
agent misunderstandings; preserve the original goal; explicitly name the app inferred 
from package\_name; keep it actionable.\\

Output (JSON only):\\
\{\\
  "refined\_high\_level\_instruction": "<refined instruction here>"\\
\}\\

\end{promptbox}

\clearpage

\phantomsection
\label{prompt:gui-ambiguity}
\clearpage

\begin{promptbox}[GUI Semantic Ambiguity Detection Prompt]
\footnotesize
You are an expert GUI interaction analyst tasked with identifying semantic-ambiguous actions in agent trajectories. Semantic-ambiguous actions are UI interactions whose functional meaning is unclear, context-dependent, or visually confounded.\\
\textbf{Task Analyze} \\
Analyze the provided trajectory and determine whether it contains any of the following three types of semantic ambiguity. A trajectory is considered to contain a type if at least one action in the sequence exhibits that characteristic.\\

\textbf{Three Types of Semantic Ambiguity}\\
\textbf{1. Context Dependency} \\
\textit{Definition}: Identical or similar UI elements that trigger different functions depending on the surrounding context, application state, or screen location.

\textit{Examples}:\\
- A "+" button that creates a new folder in one context but adds a contact in another\\
- A "Save" button that behaves differently in edit mode vs. creation mode\\
- The same icon appearing in different screens with distinct functions
\textit{Key Indicators}: Same element text/icon, different outcomes based on context; actions that require understanding the current application state or mode.\\

\textbf{2. Sequential Dependency} \\
\textit{Definition}: Actions that only succeed or make sense after completing specific prerequisite steps, or interactions that form part of a multi-step workflow where order matters critically.

\textit{Examples}:\\
- Submitting a form only after all required fields are filled\\
- Confirming a deletion only after initiating the delete action\\
- Accessing a feature that requires prior authentication or permission granting\\
- Multi-step workflows where skipping intermediate steps causes failure\\
\textit{Key Indicators}: Actions that reference “after doing X” or “following Y”; failures due to missing prerequisites; explicit workflow sequences.\\

\textbf{3. Visual Ambiguity} \\
\textit{Definition}: Visually similar or identical UI elements that correspond to distinct functions, making it difficult to distinguish their purpose based solely on appearance.

\textit{Examples}:
- Multiple icons with similar shapes but different meanings\\
- Text buttons with similar labels (“OK” vs. “Confirm” vs. “Done”) that have different effects\\
- Structurally similar list items or menu entries that perform different actions\\
- Elements that look clickable but aren’t, or vice versa\\
\textit{Key Indicators}: Reasoning mentions visual similarity, confusion between elements, or need to carefully distinguish between look-alike components.\\

\textbf{Analysis Guidelines}\\
1. Read the entire trajectory including the task goal and all reasoning-action pairs.\\
2. For each action, consider:\\
   - Does the element’s function depend on context? → Context Dependency \\
   - Does this action require prior steps to work? → Sequential Dependency \\ 
   - Are there visually similar elements that could be confused? → Visual Ambiguity \\  
3. Mark True if ANY action in the trajectory exhibits the characteristic.  \\
4. Be conservative: Only mark True if there is clear evidence in the reasoning or sequence. \\

\textbf{Input Format} \\
You will receive a trajectory string containing: \\
- Task Goal  \\
- Step-by-step reasoning-action information 

\textbf{Output Format} \\
Return ONLY a valid JSON dictionary:

\{
"context\_dependency": true,\\
"sequential\_dependency": false,\\
"visual\_ambiguity": false
\}

Important:\\
- Use lowercase with underscores for keys  \\
- Use lowercase `true` and `false` (JSON boolean format)  \\
- Do not include any explanatory text, only the JSON dictionary  \\
- All three keys must be present \\

\textbf{Trajectory to Analyze} \\
\{trajectory\_string\}

\textbf{Your Classification} \\
Provide your analysis as a JSON dictionary only:
\end{promptbox}

\clearpage
\begin{promptbox}[Prompt for Trajectory-to-Instruction Conversion (Web)]
\footnotesize
You are an expert specializing in inferring specific user tasks based on changes observed in website screenshots within an 
interaction trajectory. I will provide you with an interaction trajectory containing the following information:\\

1. Action per Step: The action performed at each step, chosen from one of these types: `click`, `scroll\_up`, `scroll\_down`, 
`input`. If the action is `input`, the input text will also be provided. Associated with each action (except scrolls) is 
information about the targeted UI element, including attributes like `bid`, `type`, `attributes`, etc.\\
2. Screenshots: A screenshot taken before each action (labeled "Step i" in the bottom-right corner, indicating it precedes 
the i-th action) and a final screenshot taken after the last action (labeled "Final" in the bottom-right). The pre-action 
screenshots will feature a green bounding box (marked with the step number 'i' in the top-left corner, also identifying 
the i-th UI element.) highlighting the element being interacted with. Pay close attention to the content within or 
associated with the green bounding box and the changes between the 'before' and 'after' screenshots for each step.\\
3. Current Page Title: The title of the current page after each action, which can provide context about the page's content or 
purpose.\\
4. Current URL: The URL of the current page after each action, which can help identify the specific page or resource being 
accessed.\\

Note: In Sub-Instruction, Analysis, Knowledge, and Task-Instruction, do not use the numerical marker (from the 
top-left of the green box) to refer to the UI element. Instead, if you need to refer to the element, use a more natural 
description based on its visual characteristics or text content (e.g., "the 'Settings' icon", "the text input field 
labeled 'Username'").\\

Your task involves two main parts:\\

Part 1: Analysis and Step Selection\\
- Analyze the entire provided trajectory (actions and screenshot changes).\\
- Identify a logically coherent subsequence of steps within the trajectory that represents a complete and reasonable user 
task.\\
- You must select actions that follow a clear and rational sequence toward achieving a specific goal.\\
- Eliminate any steps from the original trajectory that are redundant, irrelevant, irrational, or unnecessary for completing 
the identified task.\\

Part 2: Output Generation\\
Based on your analysis and step selection, devise the specific user goal or task. Your output must include five parts:\\
- Sub-Instruction (List[str]): For each step in the original trajectory, generate a natural language instruction 
corresponding to the action performed, based on the UI changes observed. This instruction should be concise, clear, 
actionable, and must incorporate key specific details visible in the screenshot, such as filenames, times, text content, 
or other relevant identifiers associated with the interacted element. Examples: "Scroll up to open the app drawer, 
revealing all installed applications.", "Click on the chat interface labeled 'General Discussion'.", "Input the 
username 'Agent' into the username field."\\
- Analysis (List[str]): For each step in the original trajectory, provide an analysis of the potential subsequent 
actions or user intent, based on the UI changes resulting from the action and its Sub-Instruction. This analysis 
should involve step-by-step reasoning, considering the observed screen state and what actions become possible or 
logical next. Example: "After tapping the '+' button, a menu with options like 'New Document' and 'New Folder' 
appeared. I can create something new. Next step would be to tap 'New Document', which might then prompt for a filename."\\
- Knowledge (List[str]): For each step `i` in the original trajectory, describe the general functionality of the UI 
element interacted with in that step. This description should be inferred by comparing the screenshot before 
action `i` (labeled "Step i") and the screenshot after action `i` (which will be the screenshot labeled "Step i+1", 
or "Final" for the last action). The description should be concise (1-2 sentences), focus on the general function 
revealed by the interaction (e.g., "Opens a settings menu," "Navigates back," "Selects an item," "Confirms an action"), 
avoid specific details unless necessary for clarity, and use generic terms like "this element" or "this button" 
or a functional description (e.g., "the search icon"). The length of this list must be equal to the number of steps in 
the original trajectory. Example: "Tapping this element initiates a search and displays matching results."\\
- Selected-Step-ID (List[int]): A list containing the integer step IDs (the 'i' from 'Step i') of the actions 
you selected in Part 1 as part of the coherent, logical task sequence. The IDs must be listed in the chronological 
order they appear in the selected subsequence.\\
- Task-Instruction (str): Based only on the selected subsequence of steps identified in Part 1, formulate a 
single, task instruction describing the overall task the user was likely trying to achieve through those 
selected steps. This instruction should correspond directly and efficiently to the selected sequence.\\

You must return only a JSON dictionary in the following format:\\
```json\\
\{\{\\
  "Sub-Instruction": List[str],\\
  "Analysis": List[str],\\
  "Knowledge": List[str],\\
  "Selected-Step-ID": List[int],\\
  "Task-Instruction": str\\
\}\}
```

The trajectory information will be provided below:\\
`{trajectory\_information}`\\

RETURN ME THE DICTIONARY I ASKED FOR.

\end{promptbox}

\clearpage
\begin{promptbox}[Web Action Reasoning \& Execution Prompt]
\footnotesize
You are an agent trying to solve a web task based on the content of the page and user instructions. You can interact 
with the page and explore, and send messages to the user. Each time you submit an action it will be sent to the browser 
and you will receive a new page.\\
\textbf{Instructions}\\
Review the current state of the page and all other information to find the best possible next action to accomplish 
your goal. Your answer will be interpreted and executed by a program, make sure to follow the formatting instructions.\\
\textbf{Goal:}\\
{goal}\\
\textbf{Observation of current step:}\\
\textbf{AXTree:}\\
Note: [bid] is the unique alpha-numeric identifier at the beginning of lines for each element in the AXTree. 
Always use bid to refer to elements in your actions.\\
Note: You can only interact with visible elements. If the "visible" tag is not
present, the element is not visible on the page.\\
{a11y\_tree}\\
\textbf{Focused element:}\\
{focused\_element}\\
\textbf{History of interaction with the task:}\\
{history}\\
\textbf{Action space:}\\
Note: This action set allows you to interact with your environment. Most of them are python function executing playwright 
code. The primary way of referring to elements in the page is through bid which are specified in your observations.\\
12 different types of actions are available:\\
noop(wait\_ms: float = 1000)\\
scroll(delta\_x: float, delta\_y: float)\\
fill(bid: str, value: str)\\
select\_option(bid: str, options: str | list[str])\\
click(bid: str, button: Literal['left', 'middle', 'right'] = 'left')\\
dblclick(bid: str, button: Literal['left', 'middle', 'right'] = 'left')\\
hover(bid: str)\\
press(bid: str, key\_comb: str)\\
focus(bid: str)\\
clear(bid: str)\\
drag\_and\_drop(from\_bid: str, to\_bid: str)\\
upload\_file(bid: str, file: str | list[str])\\
Only a single action can be provided at once. Example:\\
fill('a12', 'example with "quotes"')\\
Note:\\
- Some tasks may be game like and may require to interact with the mouse position in x, y coordinates.\\
- Some text field might have auto completion. To see it, you have to type a few characters and wait until next step.\\
- If you have to cut and paste, don't forget to select the text first.\\
- Coordinate inside an SVG are relative to it's top left corner.\\
- Make sure to use bid to identify elements when using commands.\\
- Interacting with combobox, dropdowns and auto-complete fields can be tricky, sometimes you need to use select\_option, 
while other times you need to use fill or click and wait for the reaction of the page.\\
\textbf{Abstract Example}\\
Here is an abstract version of the answer with description of the content of each tag. Make sure you follow this 
structure, but replace the content with your answer:\\
<think>\\
Think step by step. If you need to make calculations such as coordinates, write them here. Describe the effect
that your previous action had on the current content of the page.\\
</think>\\
<action>\\
One single action to be executed. You can only use one action at a time.\\
</action>\\
\textbf{Concrete Example}\\
Here is a concrete example of how to format your answer. Make sure to follow the template with proper tags:\\
<think>\\
From previous action I tried to set the value of year to "2022", using select\_option, but it doesn't appear to be in 
the form. It may be a
dynamic dropdown, I will try using click with the bid "a324" and look at the response from the page.\\
</think>\\
<action>\\
click('a324')\\
</action>\\
\end{promptbox}
\twocolumn

\end{document}